\documentclass[preprint,review,accept,moreauthors,pdftex]{Definitions/mdpi} 

\usepackage{graphicx}
\usepackage{amssymb}
\usepackage{multirow}
\usepackage{xcolor}
\usepackage{xtab}
\usepackage{longtable}
\usepackage{setspace}
\usepackage{balance}
\usepackage{caption}
\usepackage{subcaption}

\firstpage{1} 
\pubvolume{xx}
\issuenum{1}
\articlenumber{5}
\pubyear{2021}
\copyrightyear{2021}
\history{Received: date; Accepted: date; Published: date}

\Title{Machine Learning Methods for Histopathological Image Analysis: A Review}

\Author{Jonathan de Matos $^{1,2}$*\orcidA{}, Steve Tsham Mpinda Ataky $^{1}$\orcidB{}, Alceu de Souza Britto Jr. $^{2,3}$\orcidC{}, Luiz Eduardo Soares de Oliveira $^{4}$\orcidD{} and Alessandro Lameiras Koerich $^{1}$\orcidE{}
}

\AuthorNames{Firstname Lastname, Firstname Lastname and Firstname Lastname}

\address{%
$^{1}$ \quad École de Technologie Superiéure, Université du Québec, Montréal, QC, Canada; jonathan.de-matos.1@ens.etsmtl.ca, steve-tsham-mpinda.ataky.1@ens.etsmtl.ca, alessandro.koerich@etsmtl.ca\\
$^{2}$ \quad Universidade Estadual de Ponta Grossa, Ponta Grossa, PR,  Brazil\\
$^{3}$ \quad Pontifícia Universidade Católica do Paraná, Curitiba, PR, Brazil; alceu@ppgia.pucpr.br\\
$^{4}$ \quad Universidade Federal do Paraná, Curitiba, PR, Brazil; luiz.oliveira@ufpr.br}

\corres{Correspondence: jonathandematos@gmail.com}

\abstract{Histopathological images (HIs) are the gold standard for evaluating some types of tumors for cancer diagnosis. The analysis of such images is not only time and resource consuming, but also very challenging even for experienced pathologists, resulting in inter- and intra-observer disagreements. One of the ways of accelerating such an analysis is to use computer-aided diagnosis (CAD) systems. In this paper, we present a review on machine learning methods for histopathological image analysis, including shallow and deep learning methods. We also cover the most common tasks in HI analysis, such as segmentation and feature extraction. In addition, we present a list of publicly available and private datasets that have been used in HI research.}


\begin{document}

\section{Introduction}
\label{sec:intro}

\begin{figure}[htpb!]
     \centering
     \begin{subfigure}[b]{0.4\textwidth}
         \centering
         \includegraphics[width=\textwidth]{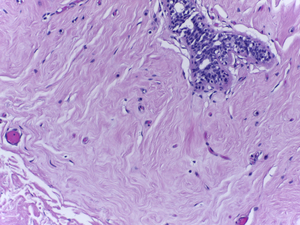}
         \caption{}
         \label{fig:y equals x}
     \end{subfigure}
     \begin{subfigure}[b]{0.4\textwidth}
         \centering
         \includegraphics[width=\textwidth]{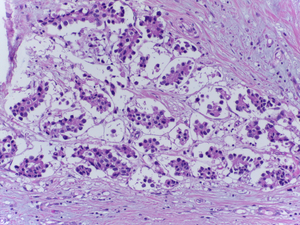}
         \caption{}
         \label{fig:three sin x}
     \end{subfigure}
     \hfill
    \caption{Example of (\textbf{a}) benign and (\textbf{a}) malignant HIs \cite{BACH2018}}
    \label{fig:adenoma_ductal_carcinoma_1}
\end{figure}

Current hardware capabilities and computing technologies provide the ability of computers to solve problems in many fields. The medical field nobly employs technologies as a means of improving populations' health and life quality. Medical computer-aided diagnosis is one of the suitable examples thereof. Amongst the aforementioned diagnosis, image-based diagnosis such as magnetic resonance imaging (MRI), X-rays, computed tomography (CT), and ultrasound have been attracting growing interest of scientists and academics. Likewise, histopathological images (HIs) are another kind of medical imaging obtained by means of microscopy of tissues from biopsies, which brings to the specialists their ability to observe tissues characteristics in a cell basis (Figure~\ref{fig:adenoma_ductal_carcinoma_1}). 

Cancer is a disease with high mortality rates in developed and in developing countries. In addition to causing death, the costs for \textcolor{black}{related} treatment are high and have an impact on the public and on the private healthcare system, penalizing, therefore, the government and the population. \textcolor{black}{According as it is mentioned} by \citet{Torre201587}, the mortality rate among high-income countries is stabilizing or even decreasing due to programs regarding the risk factors reduction (e.g. smoking, over-weighting, physical inactivity) and due to treatment improvements. In low and middle-income countries mortality rates are rising due to the increase in risk factors. One of the key points of improvements in treatment is the early detection of tumors. In fact, in 140 out of 184 countries, breast cancer is the most prevalent type of cancer among women \cite{Torre2017}. Imaging exams like mammography, ultrasound or CT can diagnose the presence of masses growing in breast tissue, notwithstanding the confirmation of which type of tumor can only be accomplished by means of a biopsy. Biopsies, in turn, take more time to provide a result due to the acquisition procedure (e.g. fine-needle aspiration or open surgical biopsy), the tissue processing (creation of slide with the staining process) and finally pathologist visual analysis. Naturally, pathologist analysis is a highly specialized and time-consuming task prone to inter and intra-observer discordance \citep{BELLOCQ2011S92}. Furthermore, the staining process can cause the variance in the process of analysis. Hematoxylin and eosin (H\&E), although \textcolor{black}{both are} the most common and accessible type of stain, they can nevertheless produce different color intensities depending on the brand, storage time, and temperature. Therefore, computer-aided diagnosis (CAD) can increase pathologists' throughput and improve the confidence of results by not only adding reproducibility to the diagnosis process but also reducing observer subjectivity.

One important feature in cancer diagnosis is the observation of nuclei. Tumors like ductal carcinoma and lobular carcinoma present an irregular growing on epithelial cells at these structures. A high number of nuclei or a high number of mitotic cells in a small region can indicate the presence of an irregular growth of tissue, representing a tumor. An HI can capture this feature, but besides the nuclei, it will capture other healthy tissues that can be seen in images of benign tumors. Stroma is a type of tissue that shows the same characteristics in parts of malignant and benign images. Selecting more relevant patches could improve the classification processes.

In the last years, we have experienced an increasing use of machine learning (ML) methods in CAD and HI analysis. ML methods have been used in the pathological diagnosis of cancer in different tissues or organs such as breast, prostate, skin, brain, bones liver, and others. ML methods have also potential advantages in HI analysis. ML methods have been widely used in segmentation, feature extraction and classification of HIs. HIs have rich geometric structures and complex textures, which are different from the visual characteristics of macro vision images used in other machine learning tasks such as object recognition, face recognition, scene reconstruction or event detection. 

In this review we attempt to capture the most relevant works of the last decade that employ ML methods for HI analysis. We present a comprehensive overview of ML methods for HI analysis including segmentation, feature extraction and classification. The motivation is to understand the development and use of ML methods in HI analysis and discover the future potential of ML methods in HI analysis. Furthermore, this review aims to address the following three research questions:
\begin{enumerate}[leftmargin=*,labelsep=4.9mm]
\item Which ML methods have been used for HI classification and how HIs are provided to the ML methods (raw images or pre-processed images or extracted features)? This question aims at identifying which monolithic classifiers, ensembles of classifiers or DL methods have been frequently used to classify HIs.
\item Which elements of HIs are considered the most important ones and how they are obtained? This question aims at identifying which types of tissues or structures can be identified using ML methods.
\item What are the trends that have been dominating HI analysis? This question aims at identifying what are the most promising ML methods for HI analysis for for the near future. 
\end{enumerate}

The main contributions of this paper are: (i) it covers a period of exponential change in the computer vision, from the handcrafted features to representation learning methods; (ii) it is a comprehensive review, which does not focus on HIs of specific tissues or organs; (iii) it categorizes the works according to the task: segmentation, feature extraction, classification, and representation learning and classification. This allows researchers to compare their works in the same context. There are several survey and review papers related to HI analysis, which are presented in Section~\ref{sec:reviews}. Different from such previous reviews or surveys on HIs that only focus on HIs of specific tissues or organs, or on a single learning modality (supervised, unsupervised or DL techniques), in this review we cover different approaches, methodologies, datasets, and experimental results, so that readers can identify possible opportunities for future research in HI analysis.

This paper is organized as follows. Section~\ref{sec:lit} proposes a taxonomy to categorize the ML methods used in HIs as well as an overview of the process of selecting journals and proceedings. Section~\ref{sec:segment} presents the segmentation methods that attempt to identify important structures in HIs, which may help to diagnosis. Section~\ref{sec:feat} presents the feature extraction methods that have been used to represent HIs for further classification. Section~\ref{sec:class} presents the shallow methods that have been used for classifying the main types of tissues and tumors in HIs. Given the importance and the growing interest in DL methods, Section~\ref{sec:deep} is devoted to present the recent approaches for HI analysis that employ such methods. Section~\ref{sec:reviews} brings together other reviews and surveys papers that have been published recently, as well as a compilation of several HI datasets that have been used in the last decade. Finally, in the last section we present the conclusions and perspectives of future works.

\section{Taxomony and Overview }
\label{sec:lit}
Based on the three research questions presented previously, we have created a search query\footnote{((histology AND image) or (histopathology AND image) or (eosin AND hematoxylin)) and (("machine learning") or ("artificial intelligence") or ("image processing"))} which was slightly adapted to each search engine. We have searched for references comprising the period of between 2008 and 2020 into five research portals (engines): IEEE Xplore, ACM Digital Library, Science Direct, Web of Science and Scopus. Table~\ref{table:1} presents the number of results obtained with the search query. We have searched based on the title, abstract and keywords for all search engines, except for Science Direct. In this case, we added the full-text search also, because the number of relevant works was very low.

\begin{table}[H]
\centering
\caption{Number of results without exclusion criteria, and after the application of the first and second exclusion criteria.}
\begin{tabular}{lrrr} 
 \toprule
\bf Search & \multicolumn{3}{c}{\bf Number of Papers} \\ 
\bf Engine & \bf Search Query & \bf After 1$^{st}$ Filter & \bf After 2$^{nd}$ Filter\\ 
 \midrule
 IEEE Xplore & 96 & 68 & -\\ 
 ACM Digital Library & 5 & 3 & -\\
 Science Direct & 1752 & 161 & -\\
 Web of Science & 409 & 54 & -\\
 Scopus & 252 & 67 & -\\
 \midrule
 \bf Total & 2514 & 353 & 178\\
 \bottomrule
\end{tabular}
\label{table:1}
\end{table}

Based on these results, the first exclusion criterion was based on the title and abstract. Most of the exclusions in this step were due to papers that mentioned "image processing" in the text, but the sense of the term was linked to the process of digitizing HIs for visual analysis by pathologists. Another exclusion criterion was the presence of the terms eosin and hematoxylin or histopathology to exclude medical images that were not the focus of this review, such as CT, MRI or radiology images. Finally, we have eliminated the duplicated articles resulting and we ended up with 353 articles. The second exclusion criterion was based on the full-text reading to evaluate the adherence of the paper's contents to the goal of this review, which has excluded almost 50\% of the papers retained by the first filtering. Therefore, we have ended up with 178 articles. Besides the papers selected using the search query, we have also included several papers related to the HI datasets used in many references cited in this review, as well as, some other references that discuss about specific ML methods and techniques that are also referred in many of the selected references.

\begin{figure}[H]
\centering
\includegraphics[width=\linewidth]{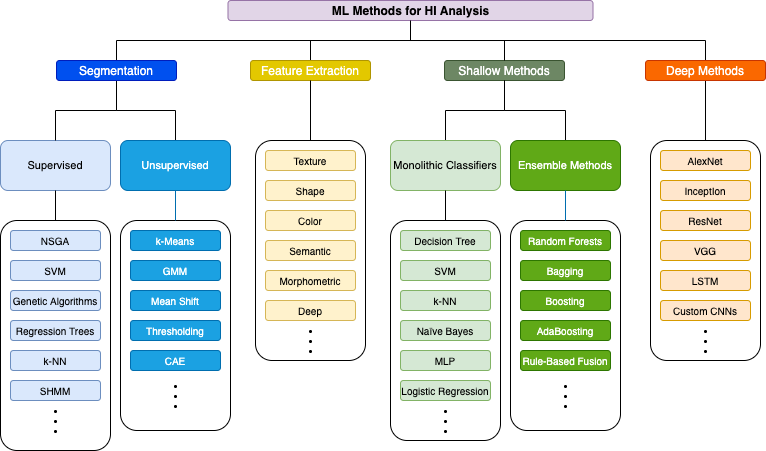}
\caption{Taxonomy used to classify HI works in this review.}
\label{fig:taxonomy}
\end{figure}

{\color{black}This review focuses on ML methods for HI analysis. Therefore, we have categorized the ML methods according to the most common ML tasks as shown in Figure~\ref{fig:taxonomy}. The top-level categories are: segmentation, feature extraction, shallow methods and deep methods. Notwithstanding DL approaches can be employed for both segmentation and classification, we proposed this division to highlight how the recent advances in DL have impacted the research on HI analysis, causing a paradigm shift to DL methods over traditional ML methods. 

Segmentation of HIs was a very popular category during the first years covered by this review. Most of the works were based on image processing techniques, such as filtering, thresholding, contour detection techniques, while others rely on ML methods, such as classification and unsupervised learning at pixel level. Besides, inasmuch as the annotation for segmentation is a very time-consuming task, it is also common to find unsupervised methods along with the supervised ones. Most of the early works used segmentation to highlighting information in HIs to specialists. 
Feature extraction aims at finding discriminative characteristics in HIs and at aggregating them into a feature vector to train ML algorithms. Most shallow classifiers and ensemble methods use such a vector representation to learn linear or non-linear decision boundaries.
We divided the category of shallow methods into two subcategories: monolithic classifiers and ensemble methods. Ensemble methods combine several diverse base models to reduce bias and/or variance in predictions as well as to improve the accuracy of predictions. The works that fall within both subcategories require a previous step of feature extraction. 

Finally, the category of deep methods contains works focused on supervised and unsupervised learning of different architectures of deep neural networks. Most of the works within this category are end-to-end learning approaches, which integrate representation learning and decision-making.}

\begin{figure}[H]
\centering
\includegraphics[width=4in]{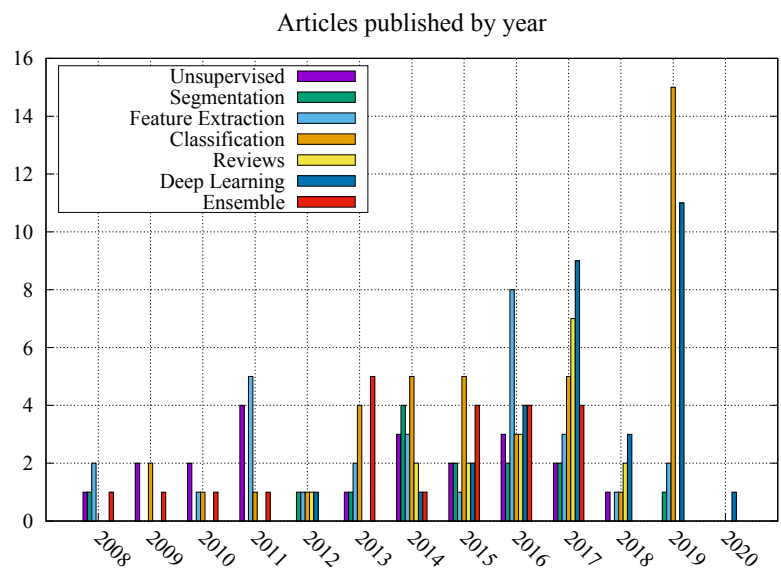}
\caption{Number of articles per year after filtering and organizing according to the main subjects.}
\label{fig:statistics}
\end{figure}

The number of publications related to the field of this research is presented in Figure~\ref{fig:statistics}. Based on Figure~\ref{fig:statistics} it is possible to note that the {research on the topic has been increasing in last years. The search was accomplished in January 2020, and regardless of the latter having been performed at the beginning of the year, some publications of the same year were found}. It is also possible to note a great increase in the use of DL methods, while ensembles and feature extraction kept their rates. Table~\ref{table:2} shows the number of publications per journal between 2008 and 2020 apropos of the subject of this review and Table~\ref{table:3} shows the top 15 journals in number of publications.

\begin{table}[H]
\centering
\caption{Top 20 journals per publication between 2008 and 2020.}
\begin{tabular}{llc}
 \toprule
 \bf Journal Title & \bf Area & \bf \# of Publications\\ 
 \midrule
Computerized Medical Imaging and Graphics & CHM & 15 \\
Medical Image Analysis & CHM & 13 \\
Pattern Recognition & C & 6 \\
Computers in Biology and Medicine & CM & 5 \\
IEEE Transactions on Medical Imaging & CEH & 5 \\
Expert Systems with Applications & CE & 4 \\
IEEE Transactions on Biomedical Engineering & E & 4 \\
Information Sciences & C & 4 \\
Applied Soft Computing & C & 3 \\
Computer Methods and Programs in Biomedicine & CM & 3 \\
Cytometry Part A & M & 3 \\
Procedia Computer Science & C & 3 \\
Artificial Intelligence in Medicine & CM & 2 \\
Computational and Structural Biotechnology Journal & BC & 2 \\
IEEE Access & CE & 2 \\
IEEE Journal of Biomedical and Health Informatics  & BCE & 2 \\
Informatics in Medicine Unlocked & M & 2 \\
Journal of Medical Imaging & M & 2 \\
Methods & MB & 2 \\
Micron & B & 2 \\
\bottomrule
\multicolumn{3}{l}{B: Biochemistry, C: Computing, E: Engineering, H: Health Sciences, M: Medicine.}\\
\end{tabular}
\label{table:2}
\end{table}

\begin{table}[H]
\centering
\caption{Top 15 conferences by number of publications between 2008 and 2020}
\begin{tabular}{lc}
\toprule
\bf Conference & \bf \# of Publications\\ 
\midrule
IEEE Intl Symp on Biomedical Imaging (ISBI) & 9 \\
IEEE Intl Conf on Healthcare Informatics, Imaging and Systems Biology & 2 \\
Intl Conf of the IEEE Engineering in Medicine and Biology Society & 2 \\
Intl Conf on Bioinformatics, Computational Biology and Health Informatics & 2 \\
Intl Conf on Information Technology in Medicine and Education (ITME) & 2 \\
Intl Conf on Pattern Recognition (ICPR) & 2 \\
Intl Symp on Medical Information Processing and Analysis & 2 \\
Medical Image Computing and Computer-Assisted Intervention & 2 \\
Medical Imaging: Digital Pathology & 2 \\
ACM Symp on Applied Computing & 1 \\
IEEE International Conference on Systems, Man, and Cybernetics (SMC) & 1 \\
IEEE Intl Conf on Bioinformatics and Biomedicine (BIBM) & 1 \\
IEEE Intl Conf on Image Processing (ICIP) & 1 \\
IEEE Intl Conf on Systems, Man, and Cybernetics (SMC) & 1 \\
IEEE Intl Symp on Multimedia & 1 \\
\bottomrule
\end{tabular}
\label{table:3}
\end{table}

\section{Segmentation Methods for HIs} \label{sec:segment}
Typically, pathologists look for tissue regions relevant to the disease being diagnosed. HI segmentation usually aims to label regions of pixels according to the structure that they may represent. For instance, the identification of nuclei structures can be used to extract morphological features, such as the number of nuclei per region, their size, and format, which may be very helpful to diagnose a tumor. In this section we present several approaches for segmenting HIs where most of them are either based on supervised or unsupervised ML methods. The former requires HI datasets with region annotation of while the latter does not require any type of annotation.

\subsection{Unsupervised Approaches}
The $k$-means algorithm is an unsupervised ML method for clustering that has been used for segmentation of pixel regions, and in the context of this review, it represents the core of fourteen segmentation methods, as shown in Table~\ref{table:unsupervised_table}.
\citet{5415659} proposed a methodology based on the expectation-maximization of the geodesic active contour for detecting lymphocyte nuclei, which is able to identify four structures: lymphocyte nuclei, stroma, cancer nuclei, and background. The process initiates with segmentation by a $k$-means algorithm, which clusters pixels of similar intensities and afterwards such clusters are improved with an expectation-maximization algorithm. The contours are identified based on the magnetic interaction theory. After contours have been defined, an algorithm searches for the concavity of contours, meaning that there is nuclei overlapping. The experiments were conducted using a breast cancer dataset. A multi-scale segmentation with $k$-means is the subject of study of 
\citet{Roullier2011603}. This work uses the same idea of the pathologist to analyze a whole slide image (WSI). The segmentation starts at a lower magnification factor and finishes at a higher magnification where it is easier to identify mitotic cells. The result of the clustering algorithm aims to identify regions of interest in each magnification. 
\citet{6061453} employed the $k$-means algorithm to help classify HIs. Although the focus is not on the $k$-means per se, but on Gabor filters, this clustering method is essential in the segmentation process. 
\citet{5872824} used $k$-means and principal component analysis (PCA) to split HIs into four types of structures: glandular lumen, stroma, epithelial-cell cytoplasm, and cell nuclei. Subsequently, morphological operations of closing and filling are performed.
\citet{6061451} used a mixture of local region-scalable fitting and $k$-means to segment cervix HIs.
\citet{ISI:000344338900003} used $k$-means for segmentation followed by skeletonization and shock graphs to identify nuclei in the previously segmented image. If the shock graph provides a confidence value smaller than 0.5 for nucleus identification, a second attempt of identification is made using a multilayer percepton (MLP). This hybrid approach achieves 92.5\% of accuracy in nucleus identification.

\citet{Mazo20161} also used $k$-means to segment cardiac images in three classes: connective tissues, light areas, and epithelial tissue. A flooding algorithm processes light areas in order to merge its result with epithelial regions and improve the final result. Finally, plurality rule was used to assign cells into flat, cubic, and cylindrical. This method achieved a sensitivity of 85\%. This work was extended in \citet{Mazo20171}.
\citet{Tosun20091104} proposed segmentation based on $k$-means that clusters all pixels into three categories (purple, pink, white), which are further divided into three subcategories. The object-level segmentation based on clustering achieved 94.89\% of accuracy against 86.78\% for pixel-level segmentation.
\citet{Nativ2014228} presented a $k$-means clustering based on morphological features of lipid droplets previously segmented using active contours models. A decision tree (DT) was used to verify the rules that lead to the classes obtained by the clustering. The correlation with pathologist evaluations reached 97\%. 
A two-step $k$-means is used by \citet{7976445} in order to segment follicular lymphoma HI. The first step segments nuclei and another type of tissues into two clusters. The next step segments "another type tissue" area from the previous step into three classes (nuclei, cytoplasm and extracellular spaces). The final step is a watershed algorithm to extract better contours of nuclei. The difference between the manual segmentation and automated was of around 1\%.
\citet{ISI:000382313300034} presented a segmentation approach based on $k$-means. The result of $k$-means segmentation is improved and simplified using a sequence of thresholds that attempt to preserve the form of objects. The key point of such a method is not the segmentation, but nucleus detection. 
\citet{Shi2017} used $k$-means to cluster pixels represented in the L*a*b color space using pixel neighborhood statistics. A thresholding step improves contours detection of fat droplets, and human specialists analyze morphological information related to the droplets to come up with a diagnosis.
\citet{Shi201799} proposed a segmentation method that considers the local correlation of each pixel. A first clustering performed by a $k$-means algorithm generates a poorly segmented cytoplasm, and a second clustering that does not consider the nuclei identified by the first clustering is performed. Finally, a watershed transform is applied to complete the segmentation.

Other clustering algorithms have also been used to segment HIs. The work proposed by \citet{Liu2008650} used the iterative self-organizing data analysis technique (ISODATA) to cluster cell images and create prototypes.
\citet{5192968} studied two strategies for initialization of clustering methods: geodesic active contours and multi-phase vector level sets. The last one proved to be more efficient when using spatial constraint fuzzy c-means, with accuracy values of 68.1\% and 67.9\% respectively, and $k$-means achieved 60.6\% in this case.
\citet{5600019} presented segmentation based on Gaussian mixture models. Their methodology uses the stain color features (hematoxylin with blue color and eosin in pink and red) to apply two segmentation steps in the red channel and other channels subsequently. It does not present ground truth comparison, only visual results compared to $k$-means.
\cite{Onder201333} presented a quasi-supervised approach based on nearest neighbors to cluster an unlabeled dataset based on itself and another labeled dataset. A comparison between quasi-supervised approach and support vector machine (SVM) has shown that SVM presents a better performance but it requires labeled data.
\citet{Yang2014996} proposed a system for content recovery based on a three-step method that uses histogram features. The first two steps use dissimilarity measures of histograms to find candidate images. The last step uses mean shift clustering. The area under the curve (AUC) of the proposed method is 0.87, which is better than 0.84 achieved by the method based on local binary patterns (LBP) features.
A mitotic cell detection system using a dictionary of cells is presented by \citet{Sirinukunwattana201516}. A shrinkage/thresholding method groups intensity features represented by a sparse coding to create a dictionary. This method achieved 80.5\% and 77.9\% of F-score on Asperio and Hamatsu subsets of MITOS dataset, respectively.
\citet{Huang2015} proposed a semi-supervised method based on exclusive component analysis (XCA) that uses the separation of stains to improve the performance. This method needs a small interaction of the user, who must provide a set of references from nuclei and from the cytoplasm.
Finally, it is worth mentioning that unsupervised methods based on DL approaches have also been proposed for segmenting HIs. We will present some recent works in Section~\ref{sec:deep}.  
\begin{table}[H]
\centering
\caption{Summary of publications on unsupervised ML methods for HI segmentation.}
\centering
\begin{tabular}{lcll} 
 \toprule
\bf Reference & \bf Year & \bf Tissue / Organ & \bf Method \\
 \midrule
\citet{Liu2008650} & 2008 & Lymph nodes & ISODATA\\
\citet{Tosun20091104} & 2009 & Colorectal & $k$-means \\
\citet{5192968} & 2009 & Prostate & Spatial constraint fuzzy $c$-means \\
\citet{5600019} & 2010 & Cervix & Gaussian mixture models\\
\citet{5415659} & 2010 & Breast & $k$-means \\
\citet{Roullier2011603} & 2011 & Breast & $k$-means \\
\citet{6061453} & 2011 & Uterus & $k$-means \\
\citet{5872824} & 2011 & Prostate & $k$-means \\
\citet{6061451} & 2011 & Uterus & $k$-means \\
\citet{Onder201333} & 2013 & Colorectal &  Quasi-supervised nearest neighbors\\
\citet{ISI:000344338900003} & 2014 & Brain & $k$-means \\
\citet{Nativ2014228} & 2014 & Liver & $k$-means \\
\citet{Yang2014996} & 2014 & Prostate & Mean shift, Similarity \\
\citet{Sirinukunwattana201516} & 2015 & Breast & Dictionary, Thresholding \\
\citet{Huang2015} & 2015 & Breast & XCA \\
\citet{Mazo20161} & 2016 & Cardiac & $k$-means \\
\citet{7976445} & 2016 & Lymph nodes & $k$-means \\
\citet{ISI:000382313300034} & 2016 & Lung & $k$-means \\
\citet{Shi2017} & 2017 & Liver & $k$-means \\
\citet{Shi201799} & 2017 & Lymph nodes & $k$-means \\
\bottomrule
\end{tabular}
\label{table:unsupervised_table}
\end{table}

\subsection{Supervised Approaches}
In this section, we present the works related to HI segmentation which are based on supervised ML approaches. Most of the works presented in this section are based on classification algorithms and therefore, they require labeled datasets in which pixels or pixel regions are annotated. Table~\ref{table:segmentation_table} summarizes the recent publications on supervised ML methods used for the segmentation purpose, where eight out of fourteen works are based on SVM classifiers.

\citet{Yu2008635} presented an approach to encode HIs using a patching procedure and a method called spatial hidden Markov model (SHMM). Each patch is represented by a feature vector that uses a mixture of Gabor energy and gray-level features. The SHMM showed improvements from 4\% to 17\% in multiple tissues in comparison to a hidden Markov model.
The work of \citet{ISI:000371029500043} uses the concept of extremal regions on gray-scale images to identify nuclei on HIs. In order to identify the threshold of extremal regions, which are organized in overlap tree, they used an SVM classifier. This approach achieved 88.5\% of F1-score against 69.8\% achieved by the state-of-the-art, considering the number of cells found after segmentation.
\citet{Janssens20131206} presented a segmentation procedure to identify muscular cells. First a segmentation based on thresholding identifies connective tissues and cells. Then, an SVM receives the segmented regions and classify them recursively into three classes (connective tissue, clump of cells and cells) until only connective and cell tissues appear. This approach achieved an F-score of 62\%, which was the state-of-the-art at that time.
\citet{Saraswat201444} proposed a segmentation procedure with a non-dominated sorted genetic algorithm (NSGA-II) and a threshold classifier. The NSGA-II generates the threshold for feature values from ground-truth images. The comparison between learned thresholds and feature values generates the segmentation.
Breast cancer prognosis is the subject of the study of \citet{6999158}. They used an SVM to perform pixel-wise classification to separate nuclei from the stroma. A second step based on a watershed algorithm identifies nuclei. The approach achieved 72\% of accuracy using pixel-level, object-level, and semantic-level features.
\citet{Salman2014295} proposed a segmentation method based on $k$-NN to analyze WSIs. The method computes histograms from patches of 64$\times$64 pixels extracted from the H\&E channels obtained by color deconvolution. The best accuracy was 73.2\% using histograms of both H\&E channels.
\citet{Chen2015} proposed a method based on pixel-wise SVM to identify stroma and tumor nests. Nuclei segmentation is carried out by a watershed algorithm, which results in 314 object-level features and 16 semantic-level features. The feature dimensionality was reduced using the analysis of feature importance.
\citet{Geessink2015} used a normal density-based quadratic discriminant classifier (QDA) to segment colorectal images. The segmentation uses the L*a*b color space with a threshold to eliminate background pixels and HSV color space to classify the remaining pixels. After classification, errors are corrected based on histological constraints. The algorithm produced an error rate of 0.6\% for tumor quantification which, according to the authors, is lower than the error of pathologists (4.4\%).
\citet{ISI:000367870000001} trained an SVM to distinguish stained pixels from unstained pixels. For such an aim, they selected manually positively stained pixels and negatively stained pixels from a set of representative images in HSV color space. The SVM identifies regions of interest for further analyses.
\citet{ISI:000380546000311} proposed an algorithm to enhance and improve general segmentation methods by utilizing a cell shape ranking function. The shape of the cells detected by the watershed transform is used to train an SVM, which discriminates real cells from false positives.
\citet{Wang20161} proposed the use wavelet decomposition, region growing, double strategy splitting model and curvature scale space to highlight nucleus regions for further classification. Textural and shape features are extracted from nuclei and feature selection is carried out based on genetic algorithms and SVM. The best results were 91.5\% and 91.6\% for sensitivity and specificity, respectively.
\citet{Arteta20163} improved the post-processing step of the method proposed by \citet{ISI:000371029500043}. Nucleus regions are refined using a surface. Two nucleus regions have their optimal area defined by a smoothness factor. The improvement provided 91\% of F1-score in the same dataset.
A nuclei segmentation was proposed by \citet{ISI:000414283200217} based on an adaptive neighborhood provided by a regression tree. A comparison showed an improvement of 9\% relative to a nuclei segmentation without adaptive thresholding.
Finally, \citet{song2019multi} presented a nuclear segmentation as a cascade of two-class classification problem. An effective learning formulation was proposed by adapting sparse convolutional models across the different layers in order to estimate the latent morphology information. For improving the region probabilities, low-level appearance and high-level contextual features from original images and probability maps estimated, respectively, are integrated into a new sequence of probabilistic binary DTs. The outcome led to a reliable contour set for each nucleus and final complete contour inferences. The experimental results over 26,500 nuclei from the Farsight, KIRC, and Kumar datasets showed that the proposed method achieved better performance than other automated segmentation approaches.
Again, it is worth mentioning that supervised methods based on DL approaches have also been proposed for segmenting HIs. We will present such recent works in Section~\ref{sec:deep}.  

\begin{table}[htpb!]
\centering
\caption{Summary of publications on supervised ML methods for HI segmentation.}
\begin{tabular}{lcll} 
 \toprule
\bf Reference & \bf Year & \bf Tissue / Organ & \bf Classifier \\
 \midrule
\citet{Yu2008635} & 2008 & Gastric & SHMM \\
\citet{ISI:000371029500043} & 2012 & Breast & Structured SVM  \\
\citet{Janssens20131206} & 2013 & Muscle & SVM \\
\citet{Saraswat201444} & 2014 & Skin & NSGA-II, Threshold \\
\citet{6999158} & 2014 & Breast & SVM \\
\citet{Salman2014295} & 2014 & Prostate & $k$-NN \\
\citet{Chen2015} & 2015 & Breast & SVM \\
\citet{Geessink2015} & 2015 & Colorectal & QDA  \\
\citet{ISI:000367870000001} & 2015 & Breast & SVM \\
\citet{ISI:000380546000311} & 2015 & Epithelium & SVM \\
\citet{Wang20161} & 2016 & Breast & GA + SVM \\
\citet{Arteta20163} & 2016 & Breast & Structured SVM \\
\citet{ISI:000414283200217} & 2017 & NA & Regression tree \\
\citet{song2019multi} & 2019 & Breast, prostate, kidney, liver, stomach, bladder & DT \\
\bottomrule
\multicolumn{4}{l}{NA: Not available.}
\end{tabular}
\label{table:segmentation_table}
\end{table}


\section{Feature Extraction for HIs}
\label{sec:feat}
Supervised shallow methods depend on the feature extraction from raw data before performing classification. HI problems require a transformation of the image pixels into meaningful features prior to classification. Feature extraction methods process images and provide a reasonable number of features summarizing the information contained in the image. In fact, feature extraction methods aim not only to reduce the dimensionality of the input, but also to highlight relevant information related to the problem (presence/absence or amount of a certain element, texture, shape, histogram, etc.) providing a representation independent on translation, scale, and rotation. {Several different types of features have been used with HIs, such as shape, size, texture, fractal, or even combination of these features.} Table~\ref{table:feature_extraction_table} summarizes the articles related to feature extraction.


Object-level and morphometric features like shape and size are particularly important for disease grading and diagnosis.
\citet{Ballaro2008703} proposed the segmentation of HIs to identify unhealthy or healthy megakaryocytes, structures from which morphometric features are extracted.
\citet{Petushi2011} employed the Otsu algorithm to highlight nuclei and then extracted different features such as inside radial contact, inside line contact, area, perimeter, area-perimeter ratio, curvature, aspect ratio, and major axis alignment. Feature vectors are built by the concatenation of histograms of all these features.   
\citet{Madabhushi2011506} presented an approach for predicting disease outcome from multiple modalities including MRI, digital pathology, and protein expression. For histopathology images, they used graph-based features such as Voronoi diagram (total area of all polygons, polygon area, polygon perimeter, polygon chord length), Delaunay triangulation (triangle side length, triangle area), minimum spanning tree (edge length), and nuclear statistics (density of nuclei, distance to nearest nuclei in different pixel radius) to represent the spatial arrangement of nuclei.
\citet{Song20131} applied thresholding and watershed transform to extract features like cystic cytoplasm length, cystic mucin production, and cystic cell density. These three features are used to train different classifiers. The experimental results showed that these three features outperformed morphological features (shape and size) achieving 90\% of accuracy against 64\%. Besides that, the combination of these features with morphological features achieved only 85\% of accuracy.
The system described by \citet{Gorelick20131804} for prostate cancer detection and classification uses a segmentation step to identify super pixels, where the segmented images are represented by morphometric and geometric features.
The framework for cytological analysis and breast cancer diagnosis presented by \citet{Filipczuk20131748} employed morphometric features. After isolation of nuclei from the images, for each nucleus they calculated area, perimeter, eccentricity, major and minor axis length, luminance mean and variance, and distance to the centroid of all nuclei.
\citet{Ozolek2014772} performed the classification of follicular lesions on thyroid tissue. After a preprocessing step for nucleus segmentation, the chromatin texture of nuclei with linear optimal transport provides features for the final classification.
\citet{Fukuma20161202} compared spatial-level and object-level descriptors like Voronoi tessellation, Delaunay triangulation, minimum spanning tree, elliptical, convex hull, bounding box and boundaries. Object-level features reached 99.07\% of accuracy at best case against 82.88\% achieved by the spatial ones.
Morphometric features can also be obtained from other structures like glands, which are easier to identify due to the difference of the lumen and other cellular structures. This is the subject in the work presented by \citet{Loeffler20121867} which uses inverse compactness and inverse solidness as measures for gland alteration on prostate cancer. The features were obtained based on the area (object and convex hull area) and perimeter of threshold highlighted objects.
\citet{Marugame2009173} used morphometric features extracted from image objects indicating nuclear aggregations to represent three categories of ductal carcinomas in breast HIs. The number of pixels, length, and thickness of the objects reflect their size and shape.
\citet{Osborne:2011:MCM:1982185.1982210} employed four geometrical features extracted from nuclei after segmentation to melanoma diagnoses in skin HIs. The four features are the ratio of the area of nuclei to the area of cytoplasm, the ratio of the perimeter of a nucleus to its area, the ratio of the major axis length of a nucleus to its minor axis length, and the ratio of the number of nuclei to the area of cytoplasm.
The multi-view approach to detect prostate cancer presented by \citet{Kwak201791} extracted morphological and intensity features from multiple resolutions. Features like area, compactness, smoothness, roundness, convex hull ratio, major-minor axis ratio, extent, bounding circle ratio, distortion, and shape context are extracted from lumens and epithelial nuclei, as well as other relational features between them.
\citet{Olgun20141390} introduced a feature extractor for HIs, which is based on the local distributions of objects, which are segmented by color intensity. The feature extractor measures the distance between an object and its neighborhood. The proposed method outperformed other 13 methods that use textural and structural features.

Texture descriptors have become quite popular in HI analysis due to the different types of textures found in HIs. For instance, high/low concentration of nuclei and stroma present quite different patterns of textures. For this reason, several researchers have been investigating a large spectrum of textural descriptors for HI classification. Descriptors based on GLCM had been used by several authors to represent textures in HI.
\citet{Kuse2010235} used GLCM as features with a pre-segmentation process based on unsupervised mean-shift clustering. Such a method reduces color variety to segment the image using thresholds. After this process nuclei are identified and have the overlapping removed by a contour and area restrictions. Finally, GLCM features are extracted from the segmented image used for classification.
\citet{Caicedo2011519} combined seven feature extraction methods, including GLCM, and create a kernel-based representation of the data on each feature type. Kernels are used inside an SVM to find similarity between data and to implement a content retrieval mechanism.
\citet{FernandezCarrobles201525} presented a feature extraction method based on frequency and spacial textons. The use of textons implies that images are represented by a reduced vocabulary of textures. Features used for the classification are histograms of textons and GLCM features extracted from texton maps. They also evaluated the impact of different colormaps on these features. Best classification results (98.1\%) were achieved by combining six color models and GLCM for textons. Despite the fact that GLCM requires a gray-level image, the conversion of the H\&E color image to gray-level is affected by the variability of the staining color, so in the end GLCM is also affected.

Another descriptor that is very often used to represent texture is the Local Binary Pattern (LBP).
\citet{Mazo20171} proposed the classification of cardiac tissues into five categories using a patching approach that aims to optimize the patch size to improve the representation. The texture of HIs was described using Local Binary Patterns (LBP), LBP Rotation Invariant (LBPri), Haralick features and different concatenations between them. Haralick features include contrast, angular second moment, homogeneity, correlation, entropy, and first and second correlation measures.
\citet{Peyret201883} applied LBP in the context of multispectral HIs. They used an SVM to evaluate the proposed LBP, which aligns all spectra and uses pixels from all other bands. It also uses a multi-scale kernel size. This feature extractor reached 99\% of accuracy compared to 88.3\% achieved by the standard LBP and 95.8\% reached by the concatenated spectra LBP.
\citet{TambascoBruno2016329} used a curvelet transform to handle multiscale HIs. The LBP algorithm extracts features from curvelets coefficients which are reduced by an ANOVA analysis.
The algorithm proposed by \citet{7532841} uses adaptive and iterative thresholding to find nuclei area and extracts texture information using LBP and histograms of oriented gradients (HOG). The proposed method achieved 93.3\% of accuracy against 92.3\% of the second-best method.
The work presented by \citet{Reis20172344} focused on the stroma maturity to evaluate breast cancer. The features for the stroma are Basic Image Features (BIF), obtained by convolving images with a bank of derivatives-of-Gaussian filters, and LBP with multiple scales for the neighborhood.
\citet{Gertych2015197} presented a system for prostate cancer classification, which also uses LBP as feature. The best accuracy was 68.4\% for cancer detection.
\citet{ISI:000383210600025} presented an invasive ductal breast carcinoma detector that extracts patches by tesselation without the square shape constraint. A set of 16,128 features derived from multiple histograms and LBP (multiple radii) using L*a*b, gray-scale and RGB color spaces are used to represent each patch.
\citet{Atupelage201361} extracted features using fractal geometry analysis, and compare them with Gabor filter bank, Leung-Malik filter bank, LBP and GLCM features. The proposed approach outperformed the other methods achieving 95\% of accuracy. 

\citet{Huang2011579} proposed a two-step feature extraction approach composed of a receptive field for detecting regions of interest and a sparse coding. The sparse coding groups features from patches of the same region. The mean and covariance matrix of receptive fields and sparse coding are the final filters. 
\citet{Noroozi2016128} proposed an automated method for discriminating basal cell carcinoma tumor from squamous cell carcinoma tumor in skin HIs using Z-transform features, which are obtained from the combination of Fourier transform features.
\citet{Wan2017291} used a dual-tree complex wavelet transform (DT-CWT) to represent the images in the context of mitosis detection in breast cancer detection. Generalized Gaussian distribution and symmetric alpha-stable distribution parameters were used as features.
\citet{ISI:000391731800013} also used fractal dimension features for breast cancer detection. These features perform well for an HI magnification of 40$\times$ to distinguish between malignant and benign tumors.

Recently, deep features have become very popular in several image classification tasks, including HIs.
\citet{KhalidKhanNiazi2016} presented a CAD system for bladder cancer that focuses on the extraction of epithelium features with segmentation using an automatic color deconvolution matrix construction.
\citet{spanhol2017} used deep features from a pre-trained AlexNet to classify breast benign and malignant tumors.
\citet{vo2019classification} presented a method for feature extraction based on the combination of CNNs and boosting tree classifiers (BTC). This method utilizes an ensemble of inception CNNs to extract visual features from multi-scale images. In the first stage, data augmentation methods were employed. Afterwards, ensembles of CNNs were trained to extract multi-context information from multi-scale images. The latter stage extracted both global and local features of breast cancer tumors.
\citet{george2019deep} proposed an approach for breast cancer diagnosis, which extracts features from nuclei based on CNNs. The methodology consists of different approaches for extracting nucleus features from HIs and select the most discriminative spatially sparse nucleus patches. A pre-trained set of CNNs was used to extract features from such patches. Subsequently, features belonging to individual images are fused using 3-norm pooling to obtain image level features. 

Finally, several works use or combine different feature categories in an attempt to capture information from both textures and geometrical structures found in HIs.
\citet{ISI:000391124500024} presented a method for quantifying instability of features across four prostate cancer datasets with known variations due to staining, preparation, and scanning platforms. They evaluated five families of features: graph-based features, which include first- and second-order descriptors of Voronoi diagrams, Delaunay triangulations, minimum spanning trees, and gland density; gland shape features, which measure the average shape of all the glands in an image, and include the lumen boundaries and the resulting area, perimeter, distance, smoothness, and Fourier descriptors; co-occurring gland tensor features, which capture the disorder of neighborhoods of glands as measured by the entropy of orientation of the major axes of glands within a local neighborhood; subgraph features, which describe the connectivity and clustering of small gland neighborhoods using gland centroids; Haralick texture features. 
\citet{ISI:000381691000001} investigated the best features for characterizing lung cancer. The authors extracted objective quantitative image features such as Haralick texture features of the nuclei (sum entropy, InfoMeas, difference variance, angular second moment), edge intensity of the nuclei, texture features of the cytoplasm and intensity distribution of the cytoplasm, Zernike shape, texture and radial distribution of intensity. 
\citet{ISI:000256869500006} proposed a low-level to high-level mapping to facilitate imaging retrieval. This mapping process consists of gray and color histograms, LBP, Tamura texture histogram, Sobel histogram, and invariant feature histograms.
\citet{Pang2017} proposed a CAD system for lung cancer detection, which uses textural features such as LBP, GLCM, and Tamura, shape features such as SIFT, global features and morphological features.
\citet{Kruk2017357} used morphometric, textural, and statistical (histogram) features to describe nuclei for clear-cell renal carcinoma grading. Genetic algorithm and Fisher discriminant were used to select the most important features.
\citet{6450064} proposed a multi field-of-view (FOV) classification scheme to recognize low versus high-grade ductal carcinoma from breast HIs. It uses a multiple patch size procedure for WSI to analyze whether morphological or textural or graph-based features is the most relevant to each patch size.
\citet{Tashk20156165} presented a complete framework for breast HI classification that estimates mitotic pixels in L*a*b color space. A combination of LBP, morphometric, and statistical features are extracted from mitotic candidates.
\citet{CruzRoa201191} proposed a patching method on HI slides to create small regions and extract scale-invariant feature transform (SIFT), luminance level, and discrete cosine transform features to create a bag-of-words. Semantic features are high-level information that can be associated with HIs to aid their classification.

\citet{5505922} compared {four color spaces (RGB, L*a*b, gray-scale and RGB)} with H\&E representation and eleven features such as Zernike, Chebychev, Chebyshev-Fourier, color histograms, GLCM, Tamura, Gabor, Haralick, edge statistics and others to represent lymph node HIs.
\citet{De2013475} propose a fusion of several feature types for uterine cervical cancer HI classification. They used a 62-dimensional feature vector based on GLCM, Delaunay triangulation and weighted density distribution.
\citet{Vanderbeck2014785} used morphological, textural and pixel neighboring statistics features to represent seven categories of white regions of liver HIs.
\citet{6868127} proposed a MIL approach to detect Barrett's cancer in HIs. They used cell-level morphometric features such as central power sums, area, radius, perimeter, and roundness of segments, maximum, mean, and minimum intensity, and intensity covariance, variance, skewness, and kurtosis within regions and patch-level features such as LBP, SIFT and color histograms from segmented images using the watershed algorithm.
\citet{6943991}\cite{7367154} proposed a feature selection method of liver HI classification based on morphometric features such as area, compactness, perimeter, aspect ratio, Zernick moment, etc., textural features such as GLCM, LBP, fractal dimension, Fourier distance, etc., and structural or graph-based features such as number of nodes/edges, modularity, pi, eta, theta, beta, alpha, gamma and Shimbel indexes, etc. Two greedy algorithms (fselector and in-house recursive) selected features in a pool of 200 features where the fitness function was implemented by an SVM classifier.
\citet{Michail20143374} highlighted nuclei using connected-component labeling to classify centroblast and non-centroblasts cells. Morphometric, textural and color features are used as features.
\citet{ISI:000403573100015} proposed the so-called geometric- and texture-aware features, which are based on Hu moments and fractal dimensional, respectively. Such a set of features was applied to detect geometrical and textural changes in nuclei to discriminate mitotic and non-mitotic cells.
The method proposed by \citet{Kong20091080} classifies neuroblastomas using textural and morphological features. It considers that pathologists use morphological features for their analysis and textural features can be easily extracted. They also use GLCM features and sequential floating forward selection to select features.

\begin{table}[htpb!]
\centering
\caption{Summary of publications devoted to feature extraction from HIs.}
\begin{tabular}{lcll} 
\toprule
\bf Reference & \bf Year & \bf Tissue / & \bf Feature \\
& & \bf Organ & \\
 \midrule
\citet{ISI:000256869500006} & 2008 & Skin & Color and gray histograms, LBP, Tamura \\
\citet{Ballaro2008703} & 2008 & Bone & Morphometric  \\

\citet{Marugame2009173} & 2009 & Breast & Morphometric  \\
\citet{Kong20091080} & 2009 & Brain & Textural, morphological\\

\citet{Kuse2010235} & 2010 & Lymph nodes & GLCM\\
\citet{5505922} & 2010 & Lymph nodes & Zernike, Chebychev, Chebyshev-Fourier, color \\ & & & histograms, GLCM, Tamura, Gabor, Haralick, \\
& & & edge statistics \\

\citet{Petushi2011} & 2011 & Breast & Morphometric  \\
\citet{Madabhushi2011506} & 2011 & Prostate & Voronoi diagram, Delaunay triangulation,\\
& & & minimum spanning tree, nuclear statistics  \\
\citet{Osborne:2011:MCM:1982185.1982210} & 2011 & Skin & Morphometric \\
\citet{Caicedo2011519} & 2011 & Skin & Gray, color, invariant feature, Sobel, Tamura\\
& & & LBP, SIFT\\
\citet{Huang2011579} & 2011 & Breast & Receptive field, sparse coding \\
\citet{CruzRoa201191} & 2011 & Skin & SIFT, luminance, DCT \\

\citet{Loeffler20121867}  & 2012 & Prostate & Morphometric  \\
\citet{Song20131} & 2013 & Pancreas & Morphometric  \\

\citet{Gorelick20131804} & 2013 & Prostate & Morphometric, geometric \\
\citet{Filipczuk20131748} & 2013 & Breast & Morphometric  \\
\citet{Atupelage201361} & 2013 & Blood & Fractal dimension  \\
\citet{6450064} & 2013 & Breast & Morphological, textural, graph-based \\
\citet{De2013475} & 2013 & Uterus & GLCM, Delaunay triangulation, weighted \\
& & & density distribution \\

\citet{Ozolek2014772} & 2014 & Thyroid & Linear optimal transport \\
\citet{Olgun20141390} & 2014 & Colorectal & Local object pattern \\
\citet{Michail20143374} & 2014 & Lymph nodes & Morphometric, texture \\
\citet{Vanderbeck2014785} & 2014 & Liver & Morphological, textural, pixel neighboring \\
& & & statistics \\
\citet{6868127} & 2014 & Esophagus & Morphometric, LBP, SIFT, color histograms \\

\citet{FernandezCarrobles201525} & 2015 & Breast & Textons \\
\citet{Gertych2015197} & 2015 & Prostate & LBP \\
\citet{Tashk20156165} & 2015 & Breast & LBP, morphometric, statistical \\
\citet{6943991}\cite{7367154} & 2015 & Liver & Morphometric, GLCM, LBP, fractal dimension, \\
& & & Graph-based \\

\citet{ISI:000383210600025} & 2016 & Breast & LBP \\
\citet{Fukuma20161202} & 2016 & Brain & Object, spatial \\
\citet{ISI:000391124500024} & 2016 & Prostate & Graph-based, shape, entropy, subgraph \\
& & & connectivity, texture\\
\citet{7532841} & 2016 & Uterus & HOG, LBP \\
\citet{TambascoBruno2016329} & 2016 & Breast & Curvelet transform, LBP \\
\citet{Noroozi2016128} & 2016 & Skin & Z-transform coefficients \\
\citet{KhalidKhanNiazi2016} & 2016 & Bladder & Morphometric  \\
\citet{ISI:000381691000001} & 2016 & Lung & Quantitative, texture \\
\citet{ISI:000391731800013} & 2016 & Breast & Fractal dimension \\

\citet{Kwak201791}  & 2017 & Prostate & Morphometric  \\
\citet{Reis20172344} & 2017 & Breast & BIF, LBP \\
\citet{Mazo20171} & 2017 & Cardiac & LBP, Haralick \\
\citet{Wan2017291} & 2017 & Breast & Wavelet transform, Gaussian distribution,\\
& & & Symmetric alpha-stable\\
\citet{spanhol2017} & 2017 & Breast & Deep \\
\citet{ISI:000403573100015} & 2017 & Oral & Hu's moment, fractal dimension, entropy\\
\citet{Pang2017} & 2017 & Lung & LBP, GLCM, Tamura, SIFT, global, morphometric\\
\citet{Kruk2017357} & 2017 & Kidney & Morphometric, textural, and statistical \\

\citet{Peyret201883} & 2018 & Prostate & LBP \\

\citet{vo2019classification} & 2019 & Breast & Deep \\
\citet{george2019deep} & 2019 & Breast & Deep  \\
\bottomrule
\end{tabular}
\label{table:feature_extraction_table}
\end{table}

In summary, given the rich geometric structures and complex textures that may be found in HIs, most of the works combine different types of features. Morphometric features are important to characterize geometric structures, but they are more complex to obtain since they require complex pre-processing, e.g. find the contour of nuclei to count them. On the other hand, textural features such as LBP and GLCM usually do not require a previous segmentation of HIs. Finally, the most recent methods of feature extraction are focused on deep features. They can be interpreted as a sequence of filters that can detect both geometric structures and textures. Therefore, deep features and deep methods seem to be very promising methods for HI analysis.


\section{Shallow Methods for HI Classification}
\label{sec:class}
ML algorithms trained in a supervised fashion can accomplish different HI analyses such as identifying types of tumors and tissues, nucleus features (e.g. mitosis phases) or specific characteristics in some organs (e.g. fat inside the liver or the size of epithelial tissue on the cervix). We present in this section the ML methods based on shallow classifiers. We start by presenting some works that employ single (monolithic) classifiers followed by classification methods based on ensemble (multiple) of classifiers. Both shallow and ensemble methods depend on a previous stage of feature extraction because they rely on handcrafted feature vectors to learn discriminant functions. Therefore, most of the feature extraction methods presented in Section~\ref{sec:feat} can be used together with the methods presented in this section.


\subsection{Monolithic Classifiers}
Different ML methods for supervised learning have been employed in HI analysis, such as support vector machines (SVM), decision trees (DT), naïve Bayes (NB), $k$-nearest neighbors ($k$-NN), multilayer perceptron (MLP), among others. Table~\ref{table:classification_table} summarizes the works reviewed in this section in terms of classification algorithm, tissue or organ from where the HI was obtained and the publication year.

SVMs are the most used classification algorithm for HIs. Several works have employed SVM with different feature categories.
\citet{Mazo20171} proposed the classification of cardiac tissues into five categories using a patching approach that aims to optimize the patch size to improve the representation. A cascade of linear SVMs separate tissues in four classes, followed by a polynomial SVM, which classifies one of these four classes in two sub-classes.
\citet{Osborne:2011:MCM:1982185.1982210} employed segmentation and morphological features with an SVM classifier to melanoma diagnoses in skin HIs. The propose approach achieved 90\% of accuracy.
\citet{Malon201297} compared the agreement of three pathologists and a ML method that uses deep features to train an SVM classifier to locate mitotic nuclei in HIs. The accuracy achieved by the SVM was 63.6\% and 98.6\% for positive and negative cases respectively, which was close to two pathologists' performance. Only one pathologist performed 99.2\% and 94.5\% on positive and negative samples, respectively.
\citet{Atupelage201361} used fractal features and an SVM to classify non-neoplastic tissues and tumors and to grade hepatocellular carcinoma HIs into five classes. The proposed approach achieved 95\% of correct classification rate for five classes and outperformed other methods that uses texture features.
\citet{Olgun20141390} introduced and approach for representation and classification of colon tissue HIs, which is based on the local distributions of objects. This approach was evaluated using an SVM and compared with other 13 methods that use textural and structural features. It outperformed all methods achieving an accuracy of 93\%.
\citet{Wan2017291} used a dual-tree complex wavelet transform (DT-CWT) to represent breast HIs for mitosis detection. Generalized Gaussian distribution and symmetric alpha-stable distribution parameters were the features used for classification with an SVM. The proposed method achieved 73\% of F-score outperforming most of other methods compared in their study.
\citet{ISI:000391731800013} used fractal features and an SVM classifier for breast cancer detection. They achieved 97.9\% of F-score for an HI magnification of 40$\times$ to distinguish between malignant and benign tumors. On the other hand, on multiclass problem, they reached a F-score of only 55.6\%.
\citet{ISI:000256869500006} proposed a low-level to high-level mapping to facilitate imaging retrieval. This mapping uses color, texture and shape features to train 18 SVMs. The experimental results compared the low-level and high-level (semantic) features, which obtained 67\% against 80\% of precision respectively, showing that the mapping from low to semantic-level features contributes favorably to the classification process.
\citet{Vanderbeck2014785} used SVM to classify white regions of liver HIs among seven classes. The best accuracy of 93.5\% was achieved for the combination of all features into a 413-dimensional feature vector. They also compared the results based on images labeled by different pathologists.
In an extension of the work of \citet{De2013475}, \citet{7307080} presented an automatic orientation detection for the epithelium with more features and used an SVM classifier.
\citet{7371235} presented a colorectal CAD system, which starts with an Otsu thresholding of red channel to separate nuclei, background, and stroma. An SVM classifier achieves 78.3\% of accuracy against 67\% of a method based on texture features.
\citet{Peikari20171078} proposed a nucleus segmentation pipeline based on multi-level thresholding and watershed algorithm on the L*a*b color space. The nucleus classification uses a cascade of SVMs. The cascade phase initially separates lymphocytes from epithelial tissue and then classify epithelial in benign and malignant. An interesting comparison with two pathologists' evaluation shows that the agreement between pathologists was 89\% and between the automated system was 74\% and 75\%. 
The classification of ovarian cancer is the subject of study of \citet{BenTaieb2017194}. The proposed method localizes regions of cancer in WSI using a multi-scale mechanism considering that each tumor type has specific characteristics which are better detected at different scales. The method automatically selects a ROI based on multiple scales. The latent variable of the latent SVM (LSVM) used for classification is the presence of a patch at a particular scale on the classification of that region. The LSVM approach achieved the accuracy of 76.2\%, which outperforms CNNs by 26\%.
\citet{Zhang201744} proposed a multi-scale classification that uses sparse coding implemented by means of Fisher discriminant analysis to construct a visual dictionary with SIFT features. The multi-scale approach using SVMs achieved the accuracy of 81.6\%, which outperformed the state-of-the-art (79.5\%).
\citet{Korkmaz20154026} proposed a classification framework based on minimum redundancy maximum relevance feature selection and least square SVM (LSSVM). They claimed the accuracy of approximately 100\% with only four false negatives for benign tumors in a three-class problem. No further comparisons were performed.

Bayesian, DT, NN, $k$-NN and other supervised ML algorithms have also been used for classification of HIs. Several works have employed such classifiers with different feature categories.
\citet{Marugame2009173} proposed a simple classifier based on Bayes theory to classify ductal carcinomas into three categories. Specialists consulted by authors claimed that the simple classifier provides, together with the morphological features, a better way to understand the results.
\citet{spanhol2017} used deep features from an AlexNet to classify breast benign and malignant tumors. The deep features were used with a logistic regression classifier. This approach achieved 90.3\% of correct classification rate for 200$\times$ magnification factor and outperformed a baseline (87.8\%) that used texture features.
\citet{De2013475} proposed grading of uterine cervical cancer using an LDA classifier. A specialist manually segmented the images to identify tumors and split them into ten segments for feature extraction. A voting strategy combined results from the segments. The best grading result was 70.5\% for the whole epithelium against 62.3\% for the vertically segmented epithelium.
\citet{Mete2009284} evaluated eleven different color spaces for representing HIs. The combination of a spherical coordinate transform and DT achieved the best accuracy, outperforming SVM and NB classifiers.
\citet{Sidiropoulos2012376} proposed a classification algorithm based on a probabilistic neural network (PNN) implemented on GPUs to grade cases of rare brain tumors. The advantage is the reduced processing time that allows an exhaustive feature combination search. For demonstration purposes, a comparison of CPU- and GPU-based algorithms showed that the GPU version takes 278 times less computation time than the CPU version for a feature vector with 20 attributes. 
The work presented by \citet{6830728} classifies follicular lymphomas using a preprocessing step to segment images based on intensity thresholds and an expectation maximization algorithm. The segmented cells are classified by LDA, achieving a detection rate of 82.6\%.
A random kitchen sink (RKS) classifier is used by \citet{ISI:000399823502195} to identify mitotic nuclei on breast cancer HIs. Nuclei are identified using thresholding in the red channel. Local active contour model selects and models nuclei. The approach achieved F-score of 96.0\% for RKS and 83.4\% for RFs on MITOS 2014 dataset.
A CAD system proposed by \citet{AngelArulJothi2016652} used HI converted to gray-scale giving priority to the red channel. Otsu thresholding guided by particle swarm optimization segments the gray-scale images and noise segments are reduced using area constraints based on the nuclei size. The closest matching rule (CMR) classifier achieved the accuracy of 100\% against 99.5\% for NB.
\citet{7455837} studied the classification of mitosis on breast cancer using a dataset labeled with the four major phases of mitosis. Classes are imbalanced, posing a challenge for the classification. They proposed a data augmentation method based on PCA and its eigenvectors and compared it to the synthetic minority over-sampling technique.
\citet{Barker201660} used a patching procedure based on a grid over the WSI to grade brain tumor type. Each patch has general features clustered using $k$-means. The final classification is performed over the features of the nuclei identified in the clustering step using elastic net model. The proposed model outperformed the methods from the 2014 MICCAI Pathology Classification challenge. 

Multiple instance learning (MIL) has also been used for classification of HIs. Several works have employed such MIL methods with different classifiers and feature categories. MIL is a weakly supervised learning paradigm that considers that instances are naturally grouped in labeled bags, without the need that all the instances of each bag have individual labels.
The MIL method proposed in \cite{6868127} to compare three MIL SVMs: SIL-SVM, MI-SVM and mi-SVM with mi-Graph. mi-Graph achieved accuracy of 87\% against 69\%  of mi-SVM. The proposed methodology is based on patching. All images are previously segmented using the watershed algorithm.
Another work from the same research group carried out a benchmark of MIL SVM methods, finding out that MILBoost gives better accuracy for instance-level approach (66.7\%) and mi-Graph performs better in bag-level prediction (72.5\%) \cite{Kandemir201544}.
A stain separation is performed in \citet{Cosatto2013} using a support vector regressor (SVR) to identify regions of interest (ROI), which is a high occurrence of hematoxylin in low-level magnification. This work uses an MIL approach because ROIs are not labeled but the WSIs, so all ROIs from a positive slide to receive positive labels. MIL uses MLP for classification, but it requires a modified loss function to represent the one-positive rule for a slide, which means that if in the prediction only one ROI appears as positive, the entire slide is positive. In the comparison between the MIL approach and SVM classification, the SVM required ROI labeling. The AUC of MIL was 0.95 against 0.94 of SVM with the advantage of reducing labeling efforts.
\citet{Xu2014591} introduced MCIL, an MIL-based method that uses patching procedure to create instance-level Gaussian classifiers which are clustered using the $k$-means algorithm. The work performs comparisons with regular image-level classification methods and MIL methods. The fully supervised method presented F-score of 76.6\% (using patch labeling) and the proposed method achieved 69.9\%. MCIL achieved 71.7\% and 60.1\% in another dataset (not patch labeled) with constrained and unconstrained MCIL respectively against 25.3\% of MIL-Boosting.
\citet{sudharshan2019multiple} compared different MIL approaches to the diagnostic of breast cancer patients. In this approach, every patient is seen as a bag, which is labeled with her diagnosis. Therefore, HIs do not need to be individually labeled, as they can share the bag label. Instances are patches extracted from the corresponding HIs, considering different magnification factors (40$\times$, 100$\times$, 200$\times$ and 400$\times$). The hypothesis is that a bag-based (patches) analysis is valuable for the analysis of HIs in comparison with single instance (entire image). The experiments were carried out on the BreakHis database using CNN, 1-NN, QDA, RF and SVM classifiers and the best accuracy of 92.1\% was achieved for 40$\times$ magnification by non-parametric MIL.

Finally, several works compare the performance of different monolithic classifiers on HIs, but without combining their predictions.
\citet{Ballaro2008703} proposed the segmentation of HIs to identify unhealthy or healthy megakaryocytes. The classification is carried out by a $k$-NN and DTs.
\citet{Song20131} trained different classifiers such as SVM, $k$-NN, neural networks, and Naïve Bayes (NB) on morphometric features. The experimental results showed that these three features outperform morphological features achieving 90\% of accuracy against 64\%. Besides that, the combination of these features with morphological features achieves only 85\% of accuracy.
\citet{TambascoBruno2016329} used a curvelet transform to handle multi-scale HIs. Texture features extracted from curvelets coefficients which are reduced by an ANOVA analysis and evaluated using DTs, SVM, RF and polynomial classifiers. They achieved an AUC of 1.00 which is higher than the best previous method (0.986).
\citet{Pang2017} proposed a CAD system for lung cancer detection. Sparse contribution analysis selects non-redundant features, which are used to train SVM, RF and extreme learning machine (ELM). Another contribution is the concave-convex variation, which consists of measuring the concavity of all nuclei in an image and use such a measurement to weight the probabilities of the classifiers. This method achieved the accuracy of 98.74\%, which is slightly better than RFs (97.68\%).
\citet{5505922} compared {four color spaces (RGB, L*a*b, gray-scale and RGB)} with H\&E representation. A weighted $k$-NN achieved the best results (99\%) followed by an RBF network and NB with 99\% and 90\% of accuracy, respectively. The best results were achieved for a color space called eosin representation.
\citet{Irshad2014390} presented a multimodal approach with multispectral images focusing on selecting the best spectral bands for classification of mitotic cells on MITOS 2012 dataset. SVM, DT, and MLP are used for classification purpose. SVM achieved the best F-score (63.7\%) using only eight best bands, which is higher than the state-of-the-art (59\%).
WSI is the core of the work proposed by \citet{Homeyer2013313}, which compares $k$-NN, NB and RFs for classification of slides based on a patching procedure and textural and intensity features. RFs with a group of all features achieved the best result (94.7\%).
\citet{khan2019health} proposed a framework for malignant cell classification in breast cytology images. Selected features train SVM, NB and RF classifiers. At the end, an ensemble method is employed to combine the classifiers based on the majority voting. The experiments have shown the accuracy of 98.0\% in the detection and classification of malignant cells.
\citet{kurmi2019microscopic} presented an approach consisting of nuclei localization in HIs and further classification as benign or malignant using MLP and SVM models. MLP achieved the best average accuracy of 95.03\%. 

\begin{table}[htpb!]
\centering
\caption{Summary of publications focusing on HI classification based on monolithic classifiers.}
\begin{tabular}{lcll} 
\toprule
\bf Reference & \bf Year & \bf Tissue / Organ & \bf Classifier \\
\midrule
\citet{ISI:000256869500006} & 2008 & Skin & SVM \\
\citet{Ballaro2008703} & 2008 & Bone & DT, $k$-NN\\

\citet{Mete2009284} & 2009 & Skin & DT, NB, SVM  \\
\citet{Marugame2009173} & 2009 & Breast & Bayes \\

\citet{5505922} & 2010 & Lymph nodes & $k$-NN, NB, RBF \\

\citet{Osborne:2011:MCM:1982185.1982210} & 2011 & Skin & SVM \\

\citet{Malon201297} & 2012 & Breast & SVM\\
\citet{Sidiropoulos2012376} & 2012 & Brain & PNN \\

\citet{De2013475} & 2013 & Uterus & LDA \\
\citet{Atupelage201361} & 2013 & Liver & SVM \\
\citet{Cosatto2013} & 2013 & Gastric & MLP (MIL) \\
\citet{Homeyer2013313} & 2013 & Liver & $k$-NN, NB, RF\\
\citet{Song20131} & 2013 & Pancreas & $k$-NN, NB, NN, SVM\\

\citet{Irshad2014390} & 2014 & Breast & DT, MLP, SVM \\
\citet{Xu2014591} & 2014 & Colorectal & Gaussian (MIL) \\
\citet{6868127} & 2014 & Gastric & SVM (MIL) \\ 
\citet{Olgun20141390} & 2014 & Colon & SVM \\
\citet{Vanderbeck2014785} & 2014 & Liver & SVM \\
\citet{6943991}~\citep{7367154} & 2014 & Liver & SVM \\
\citet{6830728} & 2014 & Lymph nodes & LDA \\

\citet{7371235} & 2015 & Colorectal & $k$-NN \\
\citet{Korkmaz20154026} & 2015 & Breast & SVM \\
\citet{Kandemir201544} & 2015 & Gastric & SVM (MIL) \\ 

\citet{ISI:000391731800013} & 2016 & Breast & SVM \\
\citet{7307080} & 2016 & Uterus & SVM\\
\citet{ISI:000399823502195} & 2016 & Breast & RKS \\
\citet{AngelArulJothi2016652} & 2016 & Thyroid & VPRS + CMR \\
\citet{Barker201660} & 2016 & Brain & Elastic net \\
\citet{TambascoBruno2016329} & 2016 & Breast & DT, Polynomial, RF, SVM \\

\citet{Pang2017} & 2017 & Liver & ELM, RF, SVM \\
\citet{Wan2017291} & 2017 & Breast & SVM \\
\citet{Mazo20171} & 2017 & Cardiac & SVM  \\
\citet{Peikari20171078} & 2017 & Breast & SVM \\
\citet{BenTaieb2017194} & 2017 & Ovary & SVM \\
\citet{Zhang201744} & 2017 & Lung & SVM \\
\citet{spanhol2017} & 2017 & Breast & Logistic regression \\

\citet{sudharshan2019multiple} & 2019 & Breast & SVM, $k$-NN, QDA, RF, CNN (MIL) \\
\citet{khan2019health} & 2019 & Breast & NB, RF, SVM \\
\citet{kurmi2019microscopic}  & 2019 & Breast & SVM, MLP \\
\bottomrule
\end{tabular}
\label{table:classification_table}
\end{table}

\subsection{Ensembles Approaches}
Ensembles approaches combine the predictions of multiple base classifiers in an attempt to improve generalization and/or robustness over a single classifier. Several researchers have proposed combining classifiers for improving the performance of HI approaches. Table~\ref{table:ensemble_table} summarizes the works reviewed in this section in terms of type of base classifier and combination strategy, tissue or organ from where the HI was obtained and the publication year.

\citet{ISI:000367870000001} employed classification using an ensemble of SVMs on ROIs segmented from WSI. Multiple "weak" classifiers trained with subsets of features and different parameters combined with a weighted sum (WS) function achieved the accuracy of 88.6\%.
\citet{Daskalakis2008196} used a preprocessing step of segmentation to enhance nuclei and extract morphological and textural features. A multi-classifier approach combines $k$-NN, linear least squares minimum distance (LLSMD), statistical quadratic Bayesian, SVM, and PNN using majority vote, minimum, maximum, average, and product rules. PNN achieved the best accuracy of 89.6\% for a base classifier while the ensemble method achieved 95.7\% with the majority vote rule.
The method proposed by \citet{Kong20091080} classifies neuroblastomas using textural and morphological features. An ensemble approach combining $k$-NN, LDA, NB and SVM classifiers using the weighted voting (WV) rule achieved the accuracy of 87.8\%.
\citet{5693834} proposed an ensemble of principal component classifiers (PCC). This ensemble classified 25 patches of each image, which are represented by 50 features. The accuracy achieved on a liver dataset was 96.41\% using the majority vote (MV) rule compared to 95.09\% achieved by a 3-NN. The same approach achieved 99.4\% of accuracy on lymphoma classification against 92.08\% achieved by the Adaboost approach.
A CAD system composed of a staining separation module, densitometric and texture feature extraction and an AdaBoost algorithm was proposed by \citet{Wang20131383}. The proposed system achieved accuracy of 94.37\% against 86.44\% of the best base classifier ($k$-NN) trained on raw H\&E images.
The system described by \citet{Gorelick20131804} uses a segmentation step to identify super pixels. An Adaboost algorithm classifies the segmented images represented by morphometric and geometric features. The system achieved the accuracy of 85\%.
A framework for cytological analysis is presented by \citet{Filipczuk20131748}. Morphometric features represent nuclei obtained after segmentation. The proposed method uses a combination of random subspaces with perceptrons to create an ensemble. The comparison showed that the ensemble approach achieved accuracy of 96.0\% compared to 90\% achieved by a boosting algorithm. \citet{ISI:000313984400007} proposed a nucleus detection method based on two Adaboost stages. The first step is based on features extracted from stain separated images. The second Adaboost refines the result of the first with line-based features. An optimal active contour refines the results from the second ensemble achieving the accuracy of 95.02\%. 
\citet{6974021} proposed an ensemble of Otsu thresholding algorithms with certain constraints and morphological operations. Four segmentation algorithms are responsible for the segmentation, but each image can have characteristics that would require different parameters for the segmentor set. The final result of segmentation is one among 18 segmentor sets that have most parameters shared with the set of segmentors that presented less difference in the segmentation. This approach achieved accuracy of 84.3\% compared to 77.4\% achieved by other methods. 

\citet{ISI:000354372500019} used an ensemble of SVM classifiers, where each model is trained with a variation of images pre-processed by Gaussian filters and color spaces. The classifiers are combined using the average rule and the best AUC value achieved was 0.978.
\citet{ISI:000380467800126} proposed a feature selection method that uses the entropy of a feature in relation to a class as a redundancy criterion and constraints in this value and the inter-feature entropy. SVM classifiers specialized in one subtype of tissue derived from prior segmentation are combined using the sum rule. The performance of ensemble approach using 37 features (94.08\%) is only 0.2\% better than the best SVM with recursive feature learning (RFE) method using 46 features. 
A comparison of multiple classifiers and features is presented by \citet{7849887}. A set of monolithic classifiers is compared with Adaboost implemented with SVM, DT, and RF. Adaboost achieved 97.8\% of accuracy.
\citet{FernandezCarrobles201699} presented a classification framework for WSI with a bagging of DTs and GLCM features which achieved 0.995 for AUC and 98.13\% for true positives.
The multi-view approach presented by \citet{Kwak201791} extracts features from multiple resolutions. A boosting algorithm combining linear SVMs and the features from multiple views achieved 0.98 of AUC compared to 0.96 of the concatenation of views.
\citet{Kruk2017357} used morphometric, textural, and statistical features to describe nuclei for classification. An ensemble made up of SVM and RF classifiers and trained with a subset of features resulting from the feature selection achieved the accuracy of 96.7\%, which was higher than the state-of-the-art (93.1\%) and the best single SVM classifier (91.1\%).
An Adaboost ensemble is used by \citet{ISI:000399332700026} to grade skin cancer. The ensemble classifies images described by features created with graph theory to represent the nuclei distribution. The ensemble achieved 72\% of accuracy.
A multi field-of-view (FOV) classification scheme is proposed by \citet{6450064}. It uses a multiple patch size procedure for WSI that firstly analyzes which features are the most relevant to each patch size. After that, it uses a RF to aggregate multiple FOV patches. They do not present a baseline for accuracy comparison, only the AUC result, showing better values for nucleus architecture features to recognize low versus high-grade ductal carcinoma.

\begin{table}[htpb!]
\centering
\caption{Summary of recent publication on ensemble approaches for HI analysis.}
\begin{tabular}{l l l l l} 
 \toprule
 \bf Reference & \bf Year & \bf Tissue / & \bf Base Classifier & \bf Combination \\
 & & \bf Organ & \bf  & \bf Rule/Function\\
 \midrule
 \citet{Daskalakis2008196} & 2008 & Thyroid & $k$-NN, LLSMD, SQ-Bayes, & Vot, Min, Max, \\
 & & & SVM, PNN & Sum, Prod \\ 
\citet{Kong20091080} & 2009 & Neuroblastoma & $k$-NN, LDA, Bayesian, SVM & WV \\
 \citet{5693834} & 2010 & Liver, & PCC & WV \\
 & & Lymphocytes & & \\ 
 \citet{DiFranco2011629} & 2011 & Prostate & SVM and RF & MV \\
 \citet{Wang20131383} & 2013 & Lung & DT  & Adaboost \\
 \citet{Gorelick20131804} & 2013 & Prostate & DT & Adaboost \\
 \citet{Filipczuk20131748} & 2013 & Breast & SVM, Perceptron & Perceptron \\
 \citet{ISI:000313984400007} & 2013 & Breast & DT, Stumps & Adaboost \\
 \citet{6450064} & 2013 & Breast & RF & MV \\
 \citet{6974021} & 2014 & Uterus & Otsu segmentors & Similarity \\
 \citet{ISI:000367870000001} & 2015 & Lymphoma & SVM & WS \\
 \citet{ISI:000354372500019} & 2015 & Prostate & SVM & Avg \\
 \citet{Gertych2015197} & 2015 & Prostate & SVM, RF & MV \\
 \citet{Tashk20156165} & 2015 & Breast & SVM, RF & MV \\
 \citet{ISI:000380467800126} & 2015 & Prostate & SVM & Sum \\
 \citet{7849887} & 2016 & Prostate & $k$-NN, SVM, DT, RF, LDA, & Adaboost \\
 & & & QDA, NB & \\
 \citet{ISI:000383210600025} & 2016 & Breast & RF & MV \\
 \citet{Wright2016125} & 2016 & Colorectal & RF & MV \\
 \citet{FernandezCarrobles201699} & 2016 & Breast & DT & Sum, Variance \\
 \citet{Kwak201791} & 2017 & Prostate & SVM & Boosting \\
 \citet{Kruk2017357} & 2017 & Kidney & SVM + RF & MV \\
 \citet{ISI:000404037600002} & 2017 & Breast & RF & MV \\
 \citet{ISI:000399332700026} & 2017 & Skin & NA & Adaboost \\ 
 \bottomrule
 \multicolumn{5}{l}{NA: Not available, WV: Weighted vote, MV: Majority vote, Avg: Average, Min: Minimum, Max: Maximum,}\\
 \multicolumn{5}{l}{Sum: Summation, Prod: Product.}\\ 
 \end{tabular}
\label{table:ensemble_table}
\end{table}

\citet{Tashk20156165} presented a complete framework for HI classification. They employ maximum likelihood estimation to obtain the mitotic pixels in L*a*b color space. A cascading classification is performed firstly with SVM and next with RFs. A comparison shows that this method achieves the accuracy of 96.5\% against 82.4\% of the best previous result in MITOS 2012 dataset.
\citet{Gertych2015197} presented a system for prostate cancer classification, which consists of SVM and RF classifiers. SVM separates the stroma and epithelium and the RF identifies benign/normal and carcinogenic tissue. The best accuracy was 68.4\% for cancer detection.
\citet{ISI:000383210600025} extended the work described in \cite{ISI:000337288300001}. The authors used simple linear iterative clustering (SLIC) to extract patches by tesselation. A set of multiple histogram and texture features are extracted from the L*a*b, gray-scale and RGB color spaces of each patch. This number of features is suitable for a RF classifier, which achieved 79.51\% of F-score for tessellated patches, compared to 77.57\% of squared patches and 71.80\% of the baseline.
SLIC is also applied by \citet{Wright2016125} in a pipeline for colorectal cancer to initially segment images. Histogram and texture features extracted from the HSV color space; \textcolor{black}{likewise, statistics from H\&E channels were extracted, in addition to GLCM as features}. A comparison showed that the proposed work achieved the {accuracy} of 79\% against 75\% from their previous work for RF. 
\citet{ISI:000404037600002} presented a system for the classification of WSI. The segmentation step uses Otsu, morphological operations and histological constraints. The classification algorithms, including RF, SVM, $k$-NN and logistic regression were trained with textural, morphometric, and statistical features extracted from random patches of segmented images. RF achieved the best accuracy (93\%).
A comparison between different ensemble approaches to classify patches of WSI is presented in \cite{DiFranco2011629}. A set of 114 features were selected and ranked using RFs. Based on the selected and ranked features, multiple linear and RBF SVMs and RF classifiers were built. The aggregation function was the majority vote. The AUCs were 0.955, 0.951 and 0.948 for RBF SVM, RF and linear SVM respectively. The best previous result was 0.935.

\section{Methods Based on Deep Learning (DL)}
\label{sec:deep}
DL methods are gaining the attention of the scientific community due to recent achievements to solve complex machine learning problems on large datasets. A convolutional neural network (CNN) is able to learn in a single optimization process, both a representation and a decision boundary. However, CNNs usually require a huge amount of data for adequate training in order to avoid overfitting problems, but most of the HI datasets have only a few patients and hundreds of images. Data augmentation \cite{MatosBOK19} \cite{Ataky2020} and transfer learning \cite{MatosBOK19a} are two possible approaches to circumvent the lack of data in HI datasets. For instance, ImageNet, which has more than 14 million images, is one of the most common datasets used for training CNNs for object recognition. Data augmentation generates new HIs from existing ones by using affine transformations or morphological operations. Another common way of data augmentation is patching HIs, which consists in producing the effect of selecting pieces of a HI with the same structure but that belong to different classes. On the other hand, the transfer learning method reuses CNNs previously trained in large datasets, which usually belongs to a different domain from the target problem. The pre-trained CNNs can be used in two ways: to extract features from HIs and use these features with shallow classifiers, as already described in Sections~\ref{sec:feat} and~\ref{sec:class}; to fine-tune such CNNs on an HI dataset, which means that filters learned on a large dataset will be adapted to the HI dataset. Recently, DL methods have been employed in HI analysis. Table~\ref{tab:deeparticles} summarizes the works reviewed in this section in terms of network architecture, tissue or organ from where the HI was obtained and the publication year.

\citet{Malon201297} were one the first authors to employ DL methods in HI analysis. They used a classical LeNet-5, a 7-layer CNN architecture proposed by \citet{726791}, which in 1998 to learn a representation from HIs previously segmented with an SVR. The features extracted by the CNN were classified by an SVM. The purpose of the classification was to find mitotic nuclei. The remarkable aspect of this work is the comparison between machines and three pathologists. The pathologists showed a Cohen Kappa factor of 0.13 and 0.44 in the best case, emphasizing the inter-observer problem.
\citet{Kainz2017} presented two CNNs based on the LeNet-5 architecture for segmentation and classification of glands in tissue of benign and malignant colorectal cancer. The first CNN separates glands from background, while the second CNN identifies gland-separating structures. Experimental results on Warwick-QU colon adenocarcinoma and GlaS@MICCAI2015 challenge datasets showed a tissue classification accuracy of 98\% and 95\%, respectively.

Some works used CNNs based on the AlexNet architecture proposed by \citet{NIPS2012_4824} in 2012. AlexNet is similar to LeNet-5 but it has 12 layers, with more filters per layer, and with stacked convolutional layers. 
\citet{7899848} employed the AlexNet CNN to classify breast cancer HIs. They compared the CNN results with some handcrafted feature extractors and shallow classifiers and they concluded that the CNN was not able to outperform the shallow methods.
\citet{7727519} evaluated architectures based on AlexNet CNN for the problem of breast cancer HI classification. The experimental results on the BreaKHis dataset showed that the CNN achieved mean accuracy rates between 81.7\% and 88.6\%, depending on the magnification, at patient level, which is better than other shallow ML approaches based on textural features.
\citet{ISI:000416616500002} also used an AlexNet CNN as well as other custom CNN architectures to classify benign and malignant tumors. Due to the small sample size, authors had to carry out data augmentation by patching and affine transforms. For cancer classification, 11 WSIs produced 231,000 images. For necrosis detection, four WSIs produced a total of 47,130 images for training. Both AlexNet and the custom CNN architectures compare favorably to most handcrafted features and a RF classifier.
\citet{budak2019computer} proposed an end-to-end model based on a pre-trained AlexNet CNN and a bidirectional LSTM (BLSTM) for detecting breast cancer in HIs. The convolutional layers are used to encode HIs into a high-level representation, which is flattened and fed into the BLSTM. Experimental results on the BreaKHis dataset showed that the proposed model achieved the best average accuracy of 96.32\% for the magnification factor of 200$\times$. Moreover, for the magnification factor of 40$\times$, 100$\times$, and 400$\times$, the average accuracy was 95.69\%, 93.61\%, and 94.29\%, respectively.

Some works use CNNs based on the inception architecture proposed by \citet{SzegedyLJSRAEVR15}. The inception modules have parallel paths where the image is passed through filters of different dimensions (1$\times$1, 3$\times$3, 5$\times$5). Additionally, max pooling is also performed. The outputs are concatenated and sent to the next inception module. GoogleLeNet, a.k.a Inception-V1 \cite{SzegedyLJSRAEVR15} has 9 such inception modules stacked linearly. It has 27 layers and employs global average pooling at the end of the last inception module. Inception-V2 and Inception-V3 \cite{SzegedyVISW15} used an upgraded inception module and auxiliary outputs, which increase the accuracy and reduce the computational complexity. Another architecture is the Inception-ResNet, which combines the inception model with the ResNet model \cite{SzegedyIVA17}. 
\citet{7493530} compared AlexNet and Inception-V1, handcrafted features and SVM, and features extracted by CNNs to classify regions of colon histology images as either gland or non-gland. The combination of handcrafted features with an SVM and the prediction of a CNN showed the best results. They used data augmentation with rotations and mirroring for handcrafted features and CNNs.
\citet{yan2019breast} integrated a pre-trained Inception-V3 with a BLSTM for classifying breast cancer HIs into normal, benign, in situ carcinoma, or invasive carcinoma. The method consists of dividing HIs into 12 small patches on average. Afterwards, a fine-tuned Inception-V3 CNN extracts features from the patches, where a 5,376-dimensional feature vector is made up of the concatenation of the weights of the last three layers of the CNN. Such feature vectors are the input of a BLSTM compounded of four layers to fuse the features of the 12 small patches and come up to an image-wise classification. The experiments show that such an approach achieved the average accuracy of 91.3\%.
\citet{MatosBOK19a} proposed a classification approach for breast cancer HIs that uses transfer learning to extract features from HIs using an Inception-V3 CNN pre-trained with the ImageNet dataset. The proposed approach improved the classification accuracy in 3.7\% using the feature extraction transfer learning and an additional 0.7\% using the irrelevant patch elimination.

Deep residual neural network (ResNet) \cite{HeZRS16} is another architecture that has been used in the classification of HIs. The residual block alleviates the problem of training very deep networks.
\citet{Khosravi2018317} evaluated the versatility of CNNs on eight different datasets of breast, lung, and bladder tissues with H\&E and immunohistochemistry images (IHC). Such an evaluation included Inception and ResNet CNNs, the combination of both CNNs, as well as the concept of transfer learning. Results showed a good performance in spite of using the raw images without any pre-processing.
\citet{vizcarra2019fusion} fused CNN and SVM outputs for HI classification. The pipeline consists of extracting SURF features for the shallow learner (SVM) and color normalization (Reinhard method) and image resizing (downsampling) for fine-tuning Inception-V3 and Inception-ResNet-V2 CNNs, pre-trained on the ImageNet dataset. CNN. The output from both shallow and deep learners are fused for final prediction. Experimental results on the BACH dataset showed a moderate accuracy of 79\% and 81\% achieved by the SVM and the CNN, respectively. On the other hand, the fusion of SVM and CNN outputs outperformed the individual learners, achieving the accuracy of 92\%.
\citet{7950667} proposed the use of a wide residual CNN to classify mitotic and non-mitotic pixels in breast HIs. The CNN is trained on mitotic and non-mitotic patches extracted from the ground truth images. Experimental results on the MICCAI TUPAC Challenge dataset showed that the wide residual CNN outperformed most of other approaches.
\citet{gandomkar2018mudern} proposed the MuDeRN framework aiming at classifying HIs into benign or malignant, and next into four subtypes. In the first stage, a ResNet with 152 layers has been trained to classify HI patches of different magnification factors as whether benign or malignant. Afterwards, the results thereof were subdivided into four subcategories of malignant and benign likewise. Lastly, for each patient, the diagnosis was conducted by combining the output of the ResNet using a meta-DT. MuDeRN achieved at the first stage the accuracy of 98.52\%, 97.90\%, 98.33\%, and 97.66\% for 40$\times$, 100$\times$, 200$\times$, and 400$\times$ magnification factors, respectively. In the second stage, MuDeRN achieved the accuracy of 95.40\%, 94.90\%, 95.70\%, and 94.60\% for 40$\times$, 100$\times$, 200$\times$, and 400$\times$ magnification factors, respectively. For patient-level diagnosis, in turn, MuDeRN achieved the accuracy of 96.25\%, considering the eight classes.
\citet{brancati2019deep} also used a ResNet to detect invasive ductal carcinoma as well as to classify lymphoma subtypes. First, the convolutional layers are trained in an unsupervised way for extracting useful features to reconstruct the input image. On the other hand, the fully connect layers are trained in a supervised way. In both cases, the softmax classifier produces a probability of the input image belonging to a given class.
\citet{talo2019automated} presented an approach based on pre-trained ResNet-50 and DenseNet-161 CNN models for automatic classification of gray-scale and color HIs. The results achieved by both CNNs outperformed the existing studies in the literature, with 95.79\% of total accuracy for the gray-scale images. ResNet-50 achieved 97.77\% of total accuracy to classify color HIs.

Another CNN architecture that has been used in HI classification is the VGG-net, which is a very uniform architecture that has 16 convolutional layers with a large number of 3$\times$3 filters. 
\citet{7950668} used a VGG-net CNN \cite{vgg2015} for classification of tissue into epithelium, stroma, and fat followed by a VGG16 CNN for classifying stroma into normal stroma or tumor-associated stroma. The first CNN achieved a pixel-level accuracy of 95.5\%, while the second CNN achieved a pixel-level binary accuracy of 92.0\%. The authors employed data augmentation by randomly rotating and flipping patches, as well as by randomly jittering the hue and saturation of pixels in the HSV color space.
The work presented by \citet{7885586} segments and distinguishes glands. They proposed an approach combining a fully convolutional network (FCN) for the foreground segmentation channel, a faster region-based CNN (R-CNN) for the object detection channel, and a holistically-nested edge detector CNN for the edge detection channel. All three CNNs are based on the VGG16 CNN. The results of these three CNNs feed another CNN that outputs a segmented image. Data augmentation by affine and elastic transformation is carried out to enhance performance and avoid overfitting. The proposed approach achieved state-of-the-art results on the dataset from the MICCAI 2015 Gland Segmentation Challenge.
\citet{kumar2020deep} developed a variant of VGG16 CNN architecture, which replaces the fully connected layers by different classifiers. The approach consists of stain normalization and data augmentation, which uses images with and without normalized stain. The augmented dataset is applied to the fused VGG16, where features are taken at the global average pooling layer. Finally, the binary classification is carried out by SVM and RF classifiers. Experiments conducted on canine mammary tumor (CMTHis) and breast cancer HIs (BreakHis), which are both randomly split into training (70\%) and test (30\%) sets. The approach achieved the accuracy of 97\%, and 93\% on BreakHis and CMTHis datasets, respectively.

Other CNN architectures have also been used in HI classification, such as DenseNet \cite{HuangLMW17} and MobileNet \cite{HowardZCKWWAA17}.
\citet{kassani2019classification} proposed an approach for classification of breast cancer HIs based on an ensemble of three pre-trained CNNs, namely VGG19, MobileNet, and DenseNet. Stain normalization, data augmentation, fine-tuning and hyper-parameter tuning were used to improve the performance of the CNNs. The multi-model ensemble method achieved better performance than single classifiers with the accuracy of 98.13\%, 95.00\%, 94.64\%, and 83.10\% for BreakHis, ICIAR, PatchCamelyon, and Bioimaging datasets, respectively.
\citet{yang2019guided} introduced the use of additional region-level supervision for classifying breast cancer HIs with a DenseNet-169 CNN pre-trained on ImageNet. For this purpose, ROIs are localized and used to guide the attention of the classification network simultaneously. This process activates neurons in regions relevant to diagnose while suppressing activation in irrelevant and noisy areas. Hence, the prediction of the network is based on the regions which a pathologist expects the network to focus on. Such an approach achieved the accuracy of 93\% on the BACH dataset. 

Finally, several works proposed custom CNN architectures for HI classification, which are usually based on some well-known architectures. The authors attempt to optimize mainly the number and the dimension of kernels and the number of layers.  
\citet{ISI:000406771302072} proposed two different CNN architectures, both with 10 layers, for breast cancer HI classification. The first CNN predicts only malignancy, while the second one predicts both malignancy and image magnification level simultaneously. Experimental results on the BreaKHis dataset showed that the magnification independent CNN approach improved the performance of magnification specific model, and that the results are comparable with previous state-of-the-art results obtained by handcrafted features. They also used data augmentation based on affine transformations.

\citet{ISI:000375550500015} introduced a CNN for aggregating annotations from crowds in conjunction with learning a model for a challenging classification task. During learning from crowd annotations phase, the CNN architecture is augmented with an aggregation layer to aggregate the ground-truth from the crowdvote matrix. Experimental results on the AMIDA13 dataset showed that the proposed CNN architecture was robust to noisy labels and positively influences the performance.
\citet{ISI:000337288300001} proposed a custom 3-layer CNN to classify patches of WSI as invasive ductal carcinoma (breast cancer) or not. Patches ended up labeled due to the region labeling. Some regions of the WSI such as background and adipose cells were excluded manually and were not patched. Patches were pre-processed using color normalization and the YUV color space. CNN outperformed an RF trained on the best handcraft feature extractor by 4\%. Compared to other works, this one has a simple protocol and uses a small network, but it was one of the precursors of CNNs to analyze HIs.
\citet{7950492} proposed an 11-layer CNN to analyze the impact of stain normalization in the training and evaluation pipeline of an automatic system for CRC tissue classification. Experimental results on the CRC dataset validated the performance of the proposed CNN as well as the role of stain normalization in CRC tissue classification.
\citet{ISI:000411791700059} proposed a 6-layer CNN to identify prostate cancer and compared it with other CNNs (AlexNet, VGG, GoogLeNet, ResNet) as well as with shallow classifiers such as SVM, RF, $k$-NN and NB. Experimental results on four tissue microarrays showed that the 6-layer CNN achieved an AUC of 0.974 and it outperformed all other approaches either based on handcrafted features with shallow classifiers or other CNN architectures.

\citet{roy2019patch} proposed five custom CNN architectures for classification of patches of breast cancer HIs. The approach consists of extracting patches, classify them and compare the result of individual patches with the one of the whole image. The output is considered correct if there is an agreement between the class labels of all extracted patches and the class label of the HI. They have also boosted the number of training samples per class using affine transformation for data augmentation. Experimental results on the ICIAR 2018 challenge dataset showed that a 14-layer CNN achieved the best results: patch-wise classification accuracy of 77.4\% and 84.7\% for four and two classes respectively; image-wise classification accuracy of 90.0\% and 92.5\% for four and two classes, respectively.
\citet{MatosBOK19} proposed a 7-layer CNN architecture based on texture filters that has fewer parameters than traditional CNNs but is able to capture the difference between malignant and benign tissues with relative accuracy. The experimental results on the BreakHis dataset showed that the proposed texture CNN achieves 85\% of accuracy for classifying benign and malignant tissues. The authors also employed data augmentation based on composed random affine transforms including flipping, rotation, and translation.
\citet{Ataky2020} proposed a novel approach for augmenting HI dataset considering the inter-patient variability by means of image blending using the Gaussian-Laplacian pyramid. Experimental results on the BreakHis dataset with a texture CNN \citep{MatosBOK19} have shown promising gains vis-à-vis the majority of DA techniques presented in the literature.
The research carried out by \citet{gecer2018detection} presented a method for breast diagnosis based on WSIs. This method aims at classifying images into five categories. At first, a salience detection was performed by a pipeline consisting of four sequential 9-layer CNNs based on the VGG-net \cite{vgg2015} architecture for multi-scale processing of the WSIs, considering different magnifications for localization of diagnostically pertinent ROIs. Afterwards, a patch-based multi-class CNN is trained on representative ROIs resulting from the consensus of three experienced pathologists. Finally, the final slide-level diagnosis is obtained by fusing the salience detector and the CNN for pixel-wise labeling of the WSIs by a majority vote rule. They claimed that the CNNs used for both detection and classification outperformed competing methods that used handcrafted features and statistical classifiers. Moreover, the proposed method achieved results comparable to the diagnoses provided by 45 pathologists on the same dataset. Experiments using 240 WSIs showed the five-class slide-level accuracy of 55\%.

\begin{table}[htpb!]
\centering
\caption{Summary of publications using DL methods in HI analysis.}
\begin{tabular}{llll} 
\toprule
\bf Reference & \bf Year & \bf Tissue / Organ & \bf Network Architecture \\
\midrule
\citet{Malon201297} & 2012 & Breast & LeNet-5 \\
\citet{ISI:000337288300001} & 2014 & Breast & 3-layer Custom\\
\citet{7899848} & 2016 & Breast & AlexNet\\
\citet{7727519} & 2016 & Breast & AlexNet \\
\citet{ISI:000406771302072} & 2016 & Breast & 10-layer Custom\\
\citet{ISI:000375550500015} & 2016 & Breast & AggNet Custom\\
\citet{7493530} & 2016 & Gland & AlexNet, Inception-V1 \\
\citet{7950667} & 2017 & Breast & Wide ResNet\\
\citet{7950668} & 2017 & Breast & VGG-net, VGG16 \\
\citet{8037745} & 2017 & Colorectal & Bilinear Custom\\
\citet{7950492} & 2017 & Colorectal & 11-layer Custom\\
\citet{Kainz2017} & 2017 & Colorectal & LeNet\\
\citet{ISI:000416616500002} & 2017 & Gastric & AlexNet, Custom\\
\citet{7885586} & 2017 & Gland & VGG16 \\
\citet{ISI:000411791700059} & 2017 & Prostate & 6-layer Custom\\
\citet{Khosravi2018317} & 2018 & Breast, Lung, Bladder & Inception-V1, ResNet \\
\citet{gandomkar2018mudern} & 2018 & Breast & ResNet\\
\citet{hou2019sparse} & 2019 & Gland, Breast & CAE+CNN Custom\\
\citet{li2019weakly} & 2019 & Breast & FCN Custom \\
\citet{vizcarra2019fusion} & 2019 & Breast & Inception-V3, Inception-ResNet-V2\\
\citet{brancati2019deep} & 2019 & Breast & ResNet \\
\citet{budak2019computer} & 2019 & Breast & AlexNet, BLSTM\\
\citet{kassani2019classification} & 2019 & Breast & VGG19, MobileNet, DenseNet \\
\citet{yang2019guided} & 2019 & Breast & DenseNet-169 \\
\citet{roy2019patch} & 2019 & Breast & 11-layer to 14-layer Custom\\
\citet{gecer2018detection} & 2019 & Breast & 9-layer Custom\\
\citet{yan2019breast} & 2019 & Breast & Inception-V3, BLSTM\\
\citet{talo2019automated} & 2019 & Breast & ResNet-50, DenseNet-161\\
\citet{kassani2019classification} & 2019 & Breast & VGG19, MobileNet, DenseNet \\
\citet{yang2019guided} & 2019 & Breast & DenseNet-169 \\
\citet{MatosBOK19} & 2019 & Breast & 7-layer Texture Custom \\
\citet{MatosBOK19a} & 2019 & Breast & Inception-V3 \\
\citet{kumar2020deep} & 2020 & Breast & VGG16\\
\citet{Ataky2020} & 2020 & Breast & 7-layer Texture Custom \\
\citet{Sheikh2020} & 2020 & Breast & 24-layer Custom \\
 \bottomrule
\end{tabular}
\label{tab:deeparticles}
\end{table}

\citet{8037745} employed a bilinear CNN (BCNN), which consists of two individual CNNs, whose outputs of the convolutional layers are multiplied with outer product at each corresponding spatial location, resulting in the quadratic number of feature maps. The input of both CNNs is H\&E images with the H and E channels separated in a pre-processing stage by a color decomposition algorithm. The proposed BCNN-based algorithm achieves the best performance with a mean classification accuracy of 92.6\%. Compared to other CNN-based algorithms, BCNN improves at least 2.4\% on classification accuracy on the CRC dataset.
\citet{li2019weakly} presented an automatic method for mitosis detection based on semantic segmentation. Such a method used a CNN in which a novel label with concentric circles was added instead of a single-pixel representation of mitosis. The inner circle represents a mitotic region whereas the ring around the inner circle is a "middle ground". This concentric loss allows training the semantic segmentation CNN with weakly annotated mitosis data. The semantic segmentation employed on breast cancer HIs to seek out mitotic cells achieved the F-score of 0.562, 0.673, 0.669, on ICPR2014, MITOSIS, AMIDA13, and TUPAC16 datasets, respectively. 
\citet{hou2019sparse} proposed a semi-supervised approach that uses a sparse convolutional autoencoder (CAE) with a crosswise constraint that decomposes patches from HIs into foreground (e.g. nuclei) and background (e.g. cytoplasm). Such a CAE initializes a supervised CNN, which carries out nucleus detection, feature extraction, and classification/segmentation in an end-to-end fashion. The experimental results not only showed that the proposed approach outperformed other approaches, but also the noteworthiness of the crosswise constraint in boosting performance. The proposed CAE-CNN achieved results comparable to the state-of-the-art using only 5\% of training data needed by other methods. 
\citet{Sheikh2020} proposed a four-input 24-layer custom CNN for classification of HIs that fuses multi-resolution hierarchical feature maps at different layers. The proposed model learns different scale image patches to account for the overall structures and texture features of cells. Experimental results on ICIAR2018 and BreaKHis datasets showed that the proposed model outperformed existing state-of-the-art models.

\begin{table}[htpb!]
\centering
\caption{Summary of the reviews and surveys on HIs and ML approaches.}
\scalebox{0.93}{
\begin{tabular}{lllll} 
 \toprule
 \bf Reference & \bf Year & \bf Image & \bf Subject & \bf Journal or Conference \\
 & & \bf Type & & \\
 \midrule
 \citet{He2012538} & 2012 & HI & Segmentation, feature & Comp Methods Progr\\ 
 & & & extraction, classification & Biomed\\
 \citet{6690201} & 2014 & HI, & Nuclei extraction, segmentation, & IEEE Reviews Biomed\\
 & & IHC & Feature extraction, classification & Eng\\
 \citet{6745404} & 2014 & HI, IHC & Segmentation & Intl Conf Electr Syst Sig\\
 & & Other & & Proc Comp Techn\\
 \citet{7192913} & 2015 & HI & Nuclei segmentation, & Intl Conf Innov Inform\\
 & & & classification & Emb Comm Sys \\
 \citet{Veta2015237} & 2015 & HI & Results of MITOS2013 Challenge & Medical Image Analysis \\
 \citet{Nawaz2016296} & 2016 & Various & Tumor ecology & Cancer Letters \\
 \citet{Madabhushi2016170} & 2016 & HI & Detection, segmentation, feature & Medical Image Analysis \\
 & & & extraction, classification & \\
 \citet{Saha2016461} & 2016 & HI & Slide preparation, staining, & Tissue and Cell\\
 & & & microscopic, imaging, & \\
 & & & preprocessing, segmentation, & \\
 & & & feature extraction, classification & \\
 \citet{Chen2017} & 2017 & HI & Image analysis of H\&E slides & Tumor Biology \\
 \citet{Robertson2017} & 2017 & Various & DL & Translat Research \\
 \citet{Cosma20171} & 2017 & HI, & Deep and shallow methods & Expert Sys App \\
 & & Other & &  \\
 \citet{AzevedoTosta201735} & 2017 & HI & Segmentation for lymphocytes & Inform Medicine Unlocked \\ 
 \citet{Litjens201760} & 2017 & MI & DL for medical images & Medical Image Analysis \\
 \citet{DiCataldo201756} & 2017 & HI & Feature extraction & Comput Struct Biotechn J\\
 \citet{Aswathy201774} & 2017 & HI & Image processing, classification & Inform Medicine Unlocked \\
 \citet{Li201866} & 2018 & MI & Content retrieval & Medical Image Analysis \\
 \citet{Komura2018} & 2018 & HI & Datasets and ML methods & Comput Struct Biotechn J \\
\citet{Zhou2020} & 2020 & HI & Classical and deep neural & IEEE Access\\
 & & & networks, classification & \\ 
\citet{Krithiga2020} & 2020 & HI & Image enhancement, segmentation, & Archives Comput \\
 & & & feature extraction, classification & Methods Eng \\
 \bottomrule
\multicolumn{5}{l}{MI: Medical images; IHC: Immunohistochemistry images.}
\end{tabular}%
}
\label{table:reviews_table}
\end{table}

\section{Reviews, Surveys and Datasets}
\label{sec:reviews}
This section brings a summary of the reviews and surveys related to HIs and ML methods. As shown in Table~\ref{table:reviews_table}, we have found {nineteen} works in this category. Reviews and surveys published between 2012 and 2015 highlight mainly approaches for nucleus segmentation and classification. On the other hand, recent publications are focused on classification of whole medical images. The reviews presented by \citet{Saha2016461}, \citet{Nawaz2016296}, \citet{Chen2017} and \citet{Robertson2017} were published in medical journals and provided a deeper view of the histology information. However, such publications overlooked aspects related to ML methods. For instance, \citet{Nawaz2016296} analyzed the characteristics of tumors and presented a brief study on how computational methods can deal with HIs. \citet{Komura2018} presented the use of ML methods in HI as well as several HI datasets. \citet{Litjens201760} reviewed DL methods applied to a variety of medical images, including HIs.
\citet{Zhou2020} presented a comprehensive overview of breast HI analysis techniques based on both classical and DL methods and publicly HI datasets. Finally, \citet{Krithiga2020} presented a systematic review of breast cancer detection, segmentation and classification on HIs focused on the performance evaluation of ML and DL techniques to predict breast cancer recurrence rates.

Given the importance of datasets for the research on HI, we have also compiled in Tables~\ref{table:datapub} and~\ref{table:datanonpub}, a list of the datasets that have been used in experiments of several works we covered in this review. We included the dataset reference, year of creation, their contents in terms of the number of images and patients, and references to the papers that have used them.

\begin{table}[H]
\centering
\caption{Summary of publicly available HI datasets.}
\label{table:datapub}
\scalebox{1}{
\begin{tabular}{llll}
\toprule
\bf Year & \bf Reference & \bf Dataset Reference &  \bf Dataset Size \\
\midrule
2010 & \citet{Kuse2010235} & \cite{icpr2010contest}  & 20 Img\\
2010 & \citet{5693834} & \cite{omegrcnia}  & 528 Img, 265 Img, 376 Img \\
2012 & \citet{ISI:000371029500043} & \cite{icpr2010contest}  & 20 Img\\
2014 & \citet{Irshad2014390} & \cite{icpr2012mitos}  & 200 Img \\
2015 & \citet{Sirinukunwattana201516} & \cite{icpr2012mitos}  & 50 WSI \\
2015 & \citet{Huang2015} & \cite{tcga001} & NA \\
2016 & \citet{ISI:000399823502195} & \cite{mitos14}  & 96 Img \\
2016 & \citet{Arteta20163} & \cite{icpr2010contest}  & 20 Img \\
2016 & \citet{ISI:000381691000001} & \cite{tcga002,tmaim}  & 2,186 WSI, 294 Img \\
2016 & \citet{ISI:000391731800013} & \cite{Spanhol2016}  & 82 Pat, 7,909 Img \\
2016 & \citet{Barker201660} & \cite{miccai14,tcga001}  & 45 Img, 604 Img \\
2016 & \citet{7849887} & \cite{tcga001}  & 682 Img \\
2017 & \citet{ISI:000403573100015} & \cite{icpr2012mitos} & 15 Img \\
2017 & \citet{Reis20172344} & \cite{prostatedataset004}  & 55 WSI \\
2017 & \citet{Mazo20171} & \cite{biscar}  & 3,000 Img \\
2017 & \citet{ISI:000404037600002} & \cite{camelyon16}  & 170 WSI, 100 WSI \\
2017 & \citet{Wan2017291} & \cite{icpr2012mitos}  & 50 Img \\
2017 & \citet{Kruk2017357} & \cite{michalkruk,michalkruk2}  & 70 Pat, 62 Pat, 94 Img \\
2018 & \citet{sudharshan2019multiple} & \cite{Spanhol2016}  & 82 Pat, 7,909 Img \\
2018 & \citet{hou2019sparse} & \cite{tcga001}, \cite{miccai14} & NA \\
2019 & \citet{gandomkar2018mudern} & \cite{Spanhol2016}  & 82 Pat, 7,909 Img \\
2019 & \citet{kumar2020deep} & \cite{Spanhol2016}, CMTHis  & 82 Pat, 7,909 Img \\
2019 & \citet{vo2019classification} & \cite{Spanhol2016}, \cite{Araujo2017} & 82 Pat, 7,909 Img, 269 Img \\
2019 & \citet{vizcarra2019fusion} & \cite{BACH2018} & 400 Img, 30 WSI \\
2019 & \citet{yan2019breast} & NA  & 249 Img \\
2019 & \citet{george2019deep} & \cite{Spanhol2016},\cite{Araujo2017}  & 82 Pat, 7,909 Img, 269 Img \\
2019 & \citet{budak2019computer} & \cite{Spanhol2016}  & 82 Pat, 7,909 Img \\
2019 & \citet{kurmi2019microscopic} & \cite{Spanhol2016} & 82 Pat, 7,909 Img \\
2019 & \citet{kassani2019classification} & \cite{Spanhol2016}, \cite{Araujo2017}, \cite{camelyon16} & 82 Pat, 7,909 Img, 269 Img \\
2019 & \citet{yang2019guided} & \cite{BACH2018}  & 400 Img, 30 WSI \\
2019 & \citet{roy2019patch} & \cite{BACH2018}  & 400 Img, 30 WSI\\
\bottomrule
\multicolumn{4}{l}{NA: Not available, Img: Images, Pat: Patients, WSI: Whole Slide Image.}
\end{tabular}}
\end{table}

\begin{table}[htpb!]
\centering
\caption{Summary of datasets that are not publicly available.}
\label{table:datanonpub}
\scalebox{1}{
\begin{tabular}{llll}
\toprule
\bf Year & \bf Reference & \bf Dataset Reference & \bf Dataset Size \\
\midrule
2008 & \citet{Yu2008635} & \cite{retrieve001} & 200 Img \\
2008 & \citet{Ballaro2008703} & NA  & 297 Img \\
2008 & \citet{Liu2008650} & NA  & 480 Img \\
2008 & \citet{ISI:000256869500006} & NA  & 1,502 Img \\
2008 & \citet{Daskalakis2008196} & NA  & 115 Img \\
2009 & \citet{Marugame2009173} & NA  & 217 WSI \\
2009 & \citet{Mete2009284} & NA  & 2 WSI \\
2009 & \citet{Kong20091080} & NA  & 389 Img \\
2009 & \citet{Tosun20091104} & NA  & 16 pat \\
2009 & \citet{5192968} & \cite{clustering001}  & 8 Img \\
2010 & \citet{5505922} & NA  & 30 WSI \\
2010 & \citet{5600019} & NA  & NA \\
2010 & \citet{5415659} & NA  & 100 Img, 9 Pat \\
2011 & \citet{Huang2011579} & NA  & 9 Slides, 36,000 Img, 40$\times$ \\
2011 & \citet{Madabhushi2011506} & \cite{prostatedataset001}, \cite{prostatedataset002}, \cite{prostatedataset003} & 58 Pat, 100 Img, 20 Pat, 40 Img, 6 Pat\\
2011 & \citet{CruzRoa201191} & NA  & 1,502 Img basal, 2,828 Img tissues \\
2011 & \citet{Caicedo2011519} & \cite{histo001}  & 6,000 \\
2011 & \citet{Petushi2011} & NA  & 30 WSI \\
2011 & \citet{Osborne:2011:MCM:1982185.1982210} & NA  & 34 cases, 126 Img \\
2011 & \citet{Roullier2011603} & NA  & NA \\
2011 & \citet{6061451} & NA  & NA \\
2011 & \citet{5872824} & NA  & 8 Pat, 62 Img \\
2011 & \citet{6061453} & NA  & 475 Img \\
2011 & \citet{DiFranco2011629} & NA  & 14 Pat, 15 Img \\
2012 & \citet{Loeffler20121867} & NA  & 125 Pat \\
2012 & \citet{Sidiropoulos2012376} & NA  & 140 cases \\
2013 & \citet{Atupelage201361} & NA  & 109 Pat, WSI 369 Img \\
2013 & \citet{Song20131} & NA  & 11 slides, 7 Pat \\
2013 & \citet{6450064} & \cite{breastboost001}, \cite{breastboost002}  & 126 Pat, 29 Pat \\
2013 & \citet{De2013475} & NA  & 62 Img \\
2013 & \citet{Homeyer2013313} & NA  & 71 Img \\
2013 & \citet{Cosatto2013} & NA  & 12,726 Pat, 12,745 WSI, 26,879 Img \\
2013 & \citet{Janssens20131206} & NA  & 111 Img \\
2013 & \citet{Onder201333} & NA  & 230 Img \\
2013 & \citet{Wang20131383} & NA  & 369 Img \\
2013 & \citet{Gorelick20131804} & NA  & 50 WSI \\
2013 & \citet{Filipczuk20131748} & NA  & 675 Img, 75 Pat \\
2013 & \citet{ISI:000313984400007} & NA  & 51 Img \\
2014 & \citet{Vanderbeck2014785} & NA  & 59 Pat \\
2014 & \citet{6868127} & NA  & 97 Pat, 214 Tissue \\
2014 & \citet{Saraswat201444} & \cite{miceleuco001}  & 30 Img \\
2014 & \citet{Olgun20141390} & NA  & 3,236 Img, 258 Pat \\
2014 & \citet{6999158} & \cite{clinical001}  & 125 Pat, 1,180 Img \\
2014 & \citet{ISI:000344338900003} & NA  & 5 Pat, 80 Img \\
2014 & \citet{Xu2014591} & \cite{weakly001}  & 10 Img, 103 Img \\
2014 & \citet{Salman2014295} & NA  & 20 Pat, 200 Img \\
2014 & \citet{Michail20143374} & NA  & 300 Img \\
2014 & \citet{Ozolek2014772} & NA  & 94 Pat \\
2014 & \citet{Nativ2014228} & NA  & 54 Img, 9 Pat \\
2014 & \citet{Yang2014996} & \cite{pleiadumdnj}  & 96 WSI \\
2014 & \citet{6974021} & NA  & 28,698 Img \\
\bottomrule
\multicolumn{4}{l}{NA: Not available, Img: Images, Pat: Patients, WSI: Whole Slide Image.}
\end{tabular}}
\end{table}

\begin{table}[H]
\centering
\begin{tabular}{llll}
\toprule
\bf Year & \bf Reference & \bf Dataset Reference  & \bf Dataset Size \\
\midrule
2015 & \citet{FernandezCarrobles201525} & NA  & 40 WSI \\
2015 & \citet{7371235} & NA  & 123 Pat, 400 Img \\
2015 & \citet{Chen2015} & NA  & 230 Pat, 1,150 Img \\
2015 & \citet{Korkmaz20154026} & NA  & 160 Img \\
2015 & \citet{ISI:000380546000311} & NA  & 350 Img \\
2015 & \citet{Tashk20156165} & \cite{icpr2012mitos}  & 50 Img \\
2015 & \citet{Kandemir201544} & \cite{oesophageal}  & 110 Img \\
2015 & \citet{ISI:000367870000001} & NA  & 101 Pat \\
2015 & \citet{Gertych2015197} & NA  & 210 Img \\
2015 & \citet{ISI:000354372500019} & \cite{DiFranco2011629}, \cite{Cheng2005} & 15 Img, 14 Pat, 9 Pat\\
2015 & \citet{ISI:000380467800126} & NA  & 40 Pat, 149 Img \\
2016 & \citet{ISI:000391124500024} & NA  & 146 WSI \\
2016 & \citet{Noroozi2016128} & NA  & 33 Img \\
2016 & \citet{Fukuma20161202} & NA  & 20 WSI \\
2016 & \citet{TambascoBruno2016329} & \cite{bioinformatics}  & 58 Img \\
2016 & \citet{Wang20161} & NA  & 68 Img \\
2016 & \citet{KhalidKhanNiazi2016} & NA  & 15 WSI, 34 Img \\
2016 & \citet{ISI:000383210600025} & NA  & 66 Img \\
2016 & \citet{Wright2016125} & \citet{colorectal004}  & 157 Pat \\
2016 & \citet{AngelArulJothi2016652} & NA  & 12 Pat, 219 Img,\\
& & & 155 Img, 64 Img \\
2016 & \citet{7532841} & NA  & 39 Pat, 390 Img \\
2016 & \citet{7976445} & NA  & 47 WSI, 423 Img \\
2016 & \citet{Mazo20161} & \cite{biscar}  & 200 Img \\
2016 & \citet{ISI:000382313300034} & NA  & 90 Img \\
2016 & \citet{FernandezCarrobles201699} & NA  & 170 WSI \\
2017 & \citet{Pang2017} & NA  & 96 WSI \\
2017 & \citet{Peikari20171078} & NA  & 121 WSI, 64 Pat \\
2017 & \citet{BenTaieb2017194} & NA  & 133 WSI \\
2017 & \citet{Zhang201744} & \cite{papsmear001}  & 285 Img, 917 Img \\
2017 & \citet{Shi2017} & NA  & 200 Img \\
2017 & \citet{Shi201799} & \cite{7976445}  & 47 WSI, 423 Img \\
2017 & \citet{Kwak201791} & NA  & 771 Img \\
2017 & \citet{ISI:000399332700026} & NA  & 907 Img, 9 WSI \\
2017 & \citet{ISI:000414283200217} & NA  & 30 WSI \\
2018 & \citet{Peyret201883} & \cite{colon001}, \cite{colon002}, \cite{colon003} & 10 Img \\
2019 & \citet{li2019weakly} & ICPR2014 MITOSIS, & NA \\
& & AMIDA13, TUPAC16 & \\
2019 & \citet{gecer2018detection} & NIH-sponsored projects  & NA \\
2019 & \citet{khan2019health} & NA & NA \\
2019 & \citet{brancati2019deep} & D-IDC & NA \\
2019 & \citet{talo2019automated} & Kimia Path24 & NA \\
\bottomrule
\multicolumn{4}{l}{NA: Not available, Img: Images, Pat: Patients, WSI: Whole Slide Image.}
\end{tabular}
\end{table}

\section{Conclusion}
\label{conclusion}
In this paper, we have presented a review of the ML methods usually employed in analysis of HIs. This review revealed an increasing interest in the classification task, while the interest in other tasks such as segmentation and feature extraction are in a clear declining in the last years, as shown in Tables~\ref{table:unsupervised_table} to~\ref{table:ensemble_table}, where the related works are arranged in ascending chronological order. We point out that the main reason for such a change is due to the introduction of DL methods, which are able to deal with raw HIs with a little or even without any pre-processing step. Normalization is one of the most used preprocessing, but in the early years other preprocessing methods such as thresholding, filtering, color models, had also been used to improve the quality of HIs for subsequent tasks such as segmentation and feature extraction, or even classification.

In the years preceding the wide adoption of DL methods, several works had focused on identifying nuclei in HIs, which are important structures to cancer diagnosis. Therefore, that lead to the exploitation of different segmentation approaches as reviewed in Section~\ref{sec:segment}. Some works used the concept of semantic features, based on the e.g. counting of nuclei, its relation to the stroma, the distance between nuclei. Stain normalization is also a recurrent topic that has appeared in several works across the years covered by this review. Such an image processing method, which reduces the color and intensity variations present in stained images, has been widely used even in conjunction with DL methods.
Feature extraction methods were the focus of interest of researchers between 2008 and 2016. Morphometric feature and textural features such as GLCM, LBP and their variants have been the most frequent features used in HI analysis, either alone or in combination with other feature types. It is important to note that the shallow classifiers require a feature extraction method. Again, the adoption of DL methods, which are able to learn representation and decision boundaries in a single optimization process, is probably the main cause of declining interest in feature extraction methods from 2016. Furthermore, pre-trained CNNs can also be used as feature extractors for HIs. Several works removed the fully connected layers of pre-trained CNNs and used the output of the last convolutional layer as feature vectors to feed shallow classifiers.
Comparing Tables~\ref{table:classification_table},~\ref{table:ensemble_table} and~\ref{tab:deeparticles} we can say that DL approaches are becoming prevalent over shallow approaches in the last five years. Although studies are still necessary for understanding how these networks learn data representation, especially with respect to HIs.

Finally, Tables~\ref{table:datapub} and~\ref{table:datanonpub} also help us to understand the increasing interest in HI analysis in the last years. We have found that most of the early works are based on small private datasets, what makes difficult for other researchers that do not have access to such HI datasets to carry out research in this area as well as to reproduce the scientific results. On the other hand, most of the recent works are based on public HI datasets, which are a great contribution to the science as they provide a way to researchers to develop new methods and compare their performance with the existing ones. However, there is still a lack of large scale supervised WSI datasets.

In conclusion, this review shows the evolution of HI analysis and the recent shift over DL methods. This review also provides valuable information to researchers in the field about datasets and other reviews and surveys.

\vspace{6pt} 



\authorcontributions{Conceptualization, J.d.M., A.S.B.Jr., L.E.S.O. and A.L.K.; Methodology, J.d.M. and S.T.M.A.; Writing--original draft preparation, J.d.M. and S.T.M.A.; Writing--review and editing, A.L.K.; Supervision, A.S.B.Jr., L.E.S.O. and A.L.K.; Funding acquisition, A.S.B.Jr. and A.L.K.; All authors have read and agreed to the published version of the manuscript.}

\funding{This research was partially funded by Natural Sciences and Engineering Research Council of Canada (NSERC) Discovery grant number RGPIN-2016-04855 and by École de Technologie Supérieure, grant Développement de Collaborations Internationales de Recherche.}

\conflictsofinterest{The authors declare no conflict of interest.} 

\abbreviations{The following abbreviations are used in this manuscript:\\

\noindent 
\begin{tabular}{@{}ll}
AUC & Area under the curve \\
CAD & Computer-aided diagnosis\\
CNN & Convolutional neural network\\
CT & Computed tomography\\
DL & Deep learning\\
DNN & Deep neural network \\
DT & Decision tree \\
ELM & Extreme learning machine\\
GLCM & Gray-level co-occurrence matrix \\
HI & Histophatologic image \\
H\&E & Hematoxylin and eosin \\
HOG & Histogram of oriented gradients\\
\end{tabular}

\noindent 
\begin{tabular}{@{}ll}
IHC & Immunohistochemistry images \\
Img & Images \\
LBP & Local binary patterns \\
ML & Machine learning\\
MIL & Multiple instance learning \\
MLP & Multilayer perceptron\\
MRI & Magnetic resonance imaging \\
NSGA & Non-dominated sorted genetic algorithm \\
Pat & Patients \\
PCA & Principal component analysis \\
RCNN & Recurrent convolutional neural network \\
RF & Random forest \\
ROI & Region of interest \\
SHMM & Spatial hidden Markov model\\
SIFT & Scale-invariant feature transform \\
SNN & Synergistic neural network\\
SVM & Support vector machine \\
WSI & Whole slide image \\
XCA & Exclusive component analysis \\
\end{tabular}}


\reftitle{References}


\externalbibliography{yes}
\bibliography{sample.bib}


\end{document}